\PassOptionsToPackage{table,dvipsnames}{xcolor}
\documentclass[accepted]{uai2025} % after acceptance, for a revised version; 
\usepackage[american]{babel}
\usepackage{natbib} % has a nice set of citation styles and commands
\usepackage{mathtools} % amsmath with fixes and additions
\usepackage{booktabs} % commands to create good-looking tables
\usepackage{tikz} % nice language for creating drawings and diagrams
\usetikzlibrary{positioning}

\usepackage{amssymb}
\usepackage{amsmath}

\usepackage{subcaption}
\usepackage{physics}
\usepackage{graphicx}

\newtheorem{proposition}{Proposition}
\newtheorem{definition}{Definition}
\newtheorem{corollary}{Corollary}

\DeclarePairedDelimiter{\ceil}{\lceil}{\rceil}

\newcommand{\zonotope}{\mathcal{Z}}
\newcommand{\nestedzonotope}{\mathcal{Z}^{\alpha}_{p_Z}}

\title{Guaranteed prediction sets for functional surrogate models}

\author[1]{Ander~Gray}
\author[2,3]{Vignesh~Gopakumar}
\author[1]{Sylvain~Rousseau}
\author[1]{S\'ebastien~Destercke}
\affil[1]{%
    Universit\'e de technologie de Compi\`egne\\
    CNRS\\
    Heudiasyc\\
    France
}
\affil[2]{%
    UK Atomic Energy Authority
}
\affil[3]{%
    University College London\\
    UK
  }
  
\begin{document}
\maketitle

\begin{abstract}
We propose a method for obtaining statistically guaranteed prediction sets for functional machine learning methods: surrogate models which map between function spaces, motivated by the need to build reliable PDE emulators. The method constructs nested prediction sets on a low-dimensional representation (an SVD) of the surrogate model's error, and then maps these sets to the prediction space using set-propagation techniques. This results in prediction sets for functional surrogate models with conformal prediction coverage guarantees. We use zonotopes as basis of the set construction, which allow an exact linear propagation and are closed under Cartesian products, making them well-suited to this high-dimensional problem. The method is model agnostic and can thus be applied to complex Sci-ML models, including Neural Operators, but also in simpler settings. We also introduce a technique to capture the truncation error of the SVD, preserving the guarantees of the method.
\end{abstract}

\section{Introduction}
One struggles to find an engineering or scientific discipline which has remained untouched from the rapid progression in machine learning (ML) and artificial intelligence (AI). Scientific machine learning (Sci-ML), with tailored architectures for physics-based problems, have in particular driven major advancements. Neural Operators (NOs) \citep{li2020fourier}, neural networks which can learn between function spaces, have received attention due to their efficient surrogacy of partial differential equation (PDE) solvers \citep{azizzadenesheli2024neural}, and have seen application in complex problems including weather modelling \citep{kurth2023fourcastnet} and plasma physics \citep{gopakumar2024plasma}.

However, the wide application of AI methods continues despite widely-shared concerns, as adumbrated by~\cite{brundage2020toward},~\cite{dalrymple2024towards}, and many others, about the reliability and soundness of these methods. This is an opinion that we share: that methods for AI reliability are underdeveloped in comparison to its progression and wide application.

Although multiple methods exist for uncertainty analysis in AI, little R\&D effort has gone into approaches which can provide \textit{quantitative safety guarantees} on the predictions of AI systems. Probabilistic machine learning methods, such as Bayesian neural networks, Gaussian processes, Monte Carlo drop-out, and deep ensembles, are powerful methods which can equip predictions with distributional uncertainties. These include many notable works in neural PDE surrogates and operator learning, including~\citep{YANG2022115399, yang2019adversarial, beltran2024galerkin}

However, they do not attempt to provide statistical guarantees. By \textit{statistical guarantee} we refer to a provable property of the model’s uncertainty, given some quite weak assumptions about the randomness of the data. In this paper, we pursue a method based on conformal prediction~\citep{vovk2005algorithmic,shafer_CP}, which is less committal than purely probabilistic approaches, but can yield such guarantees. Rather than producing a full predictive distribution, the method gives a \textit{set-valued} prediction $\mathbb{C}^{\alpha}$ equipped with a confidence level $1-\alpha$. This prediction set can be \textit{guaranteed} to contain the next unobserved true label $Y_{n+1}$, with at least probability $1-\alpha$:
\begin{equation*}
  \mathbb{P}(Y_{n+1} \in \mathbb{C}^{\alpha}) \geq 1 - \alpha.
\end{equation*}
The prediction set $\mathbb{C}^{\alpha}$ is constructed using previously observed data $(X_i, Y_i)$, and the guarantee holds if the sequence $(X_1, Y_1), \ldots, (X_n, Y_n), (X_{n+1}, Y_{n+1})$ is exchangeable, e.g., if the new data point is drawn from the same joint distribution.

This paper's goal is to extend this framework to models with function-valued outputs. That is, given an input $X$, we want to predict not a single scalar, but a full function $F$ (large vector or tensor e.g., a time series, a spatial field, or a parameterized curve). Our goal is to construct a set of functions that is guaranteed, with a user-defined probability, to contain the true function associated with a new input. Our primary motivation for this is to develop reliable uncertainty quantification for PDE surrogate models, where the functional data represents solutions from numerical PDE solvers. However, the method could be faithfully applied to non-physics cases. Unlike scalar predictions, functional data can exhibit complex dependencies across their domain (e.g., neighboring time points or spatial locations), which our method must capture to produce meaningful predictions. Finally, we want this guarantee to hold using only a held-out calibration dataset. The following formal setup further describes this problem.
%They can show asymptotically correct results, meaning that in the as data availability is increased, the posterior distribution's credible intervals do approach the correct coverage probability. Likely, at least in part, due to the fact that the impact of the prior is reduced as more data is made available in Bayesian inference. 
% It has also recently been shown by~\cite{balch2019satellite} that Bayesian posteriors (even exact solutions) can suffer from \textit{False Confidence}, meaning that by their nature (additivity) they can produce unsafe results, including in practical/real-world risk calculations, as they demonstrate. Although the advancements in Bayesian machine learning methods has improved AI reliability, for the above reasons alternatives should perhaps be sought for safety-critical systems.
\paragraph{Problem statement} Given a pre-trained model $\hat{f}: X \mapsto F$, which maps to a space of functions $F \in \mathcal{F}$, and some additional (calibration) data unseen by the model $Z = (Z_1, \dots, Z_n)$ where $Z_i = (X_i, F_i)$, construct a prediction set $\mathbb{C}^{\alpha} \subset \mathcal{F}$ (a set of functions) guaranteed to enclose a next unseen observation $F_{n+1}$ with a user prescribed confidence level: $\mathbb{P}(F_{n+1} \in \mathbb{C}^{\alpha}) \geq 1 - \alpha$. The space $\mathcal{F}$ has been discretised $F_1 = [F_1(y_1), \dots,F_1(y_l)] \in \mathbb{R}^l$, but rather finely $l \gg 1$.
% \paragraph{Challenge} What are the challenges for achieving the problem statement? High dimensionality, multivariate prediction sets, input dependent predictions. The field predictions may also be highly correlated.
\subsection{Summary of the method}\label{sec:steps}
We briefly outline our strategy.
\begin{enumerate}
  \item Compute $\hat{f}$'s error with respect to the calibration data $e_i = F_i - \hat{f}(X_i)$,% with $e \in \mathbb{R}^{n\times m}$
  \item Perform a dimension reduction (e.g. an SVD) of e, and project it to a lower dimensional space,
  % an SVD of the error $e = U \Sigma V^{\top}$, and truncate dimensions, %$U \in \mathbb{R}^{n\times k}$, with $k < m$
  \item Find an enclosing set $\zonotope$ of the dimension reduced error, $U_i \in \zonotope$ for all $i$, and a point $p_Z$ which is close to the error's mode. In this work we use zonotopes for $\zonotope$,
  \item Construct nested prediction regions $\mathcal{Z}^{\alpha}$ using $\zonotope$ and $p_Z$, such that  $\mathbb{P}(U_{n+1} \in \mathcal{Z}^{\alpha}) \geq 1 - \alpha$,
  \item Bound truncation error by taking the Cartesian product of the prediction regions $\mathcal{Z}^{\alpha}$ and a bounding box $E$ of the data of the truncated dimension, $\mathcal{R}^{\alpha} = \mathcal{Z}^{\alpha} \times E$,
  \item Project $\mathcal{R}^{\alpha}$ back, and add to the result of the model's prediction $\mathbb{C}^{\alpha} = \hat{f}(X_{n+1}) + \mathcal{R}^{\alpha}$.
\end{enumerate}

\subsection{Related work}
While we review several prior methods, our coverage may not be exhaustive due to the rapidly evolving literature, since uncertainty quantification in Sci-ML is quite a timely problem.

\paragraph{Copula-based conformal prediction:}~\cite{messoudi2021copula} suggest a method to combine univariate prediction sets obtained from conformal prediction using \textit{copulas}, powerful aggregation functions used to decompose multivariate distributions into their marginals and dependencies. This work was further extended by \cite{sun2023copula} to time series. We note however that using copulas to model dependence in high dimensions is challenging, and their proposition may require advanced copula methods, such as vine-copulas, to be applicable to functional surrogates.
\paragraph{Supremum-based conformal prediction:} a quite straightforward way to combine univariate conformal predictors is by taking the supremum over a collection of non-conformity scores. \cite{diquigiovanni2022conformal} take this idea forward by proposing that the scores of each dimension can be normalised or \textit{modulated} by some function $\sigma(t)$:
\begin{equation*}
  s(x, y) = \sup_{i \in 1,\dots,N}\left(\sup_{t \in \mathcal{T}} \Big| \frac{y_i(t) - \mu(t)}{\sigma(t)} \Big|\right),
\end{equation*}
in some prediction region $\mathcal{T}$ of interest, with $\mu(t)$ being the estimated data mean. They suggest that 
\begin{equation*}
  \sigma(t) = \sqrt{\frac{\sum^{N}_{i=1} (y_i(t) - \mu(t))^2}{N - 1}} 
\end{equation*}
is taken to be the estimated data standard deviation. Their method is quite easily adapted to surrogate modelling, by replacing the $\mu(t)$ with a trained predictor $\hat{f}(x_i)(t)$ in both of above expressions, with $\sigma(t)$ now giving the standard deviation of $\hat{f}$'s prediction error. Although this method is simple to apply, we find it can at times give quite wide prediction sets.
\paragraph{Quantile-functional conformal prediction:}~\cite{ma2024calibrated} propose a quantile neural operator, which they calibrate to give a PAC-style bound on the percentage of points in the function domain that falls within a predicted functional uncertainty set. Our methods diverge as we aim to guarantee the entire function $\mathbb{P}(F_{n+1} \in \mathbb{C}^{\alpha}) \geq 1 - \alpha$ for all values of $F_{n+1}$ simultaneously. While they control the proportion of the function domain covered (with respect to a uniform sampling of $F_{n+1}$'s domain), we aim for full-function coverage. Our method also differs as it does not require a quantile function (an additional neural operator, with additional training data) to be trained (but we do require SVDs).
\paragraph{Elliptical-set conformal prediction:}~\cite{messoudi2022ellipsoidal} propose a multi-target (multivariate) non-conformity score
\begin{equation*}
  s(x,y) = \sqrt{(y - \hat{f}(x))^{\top}\hat{\Sigma}^{-1}(y - \hat{f}(x)) } ,
\end{equation*}
where $\hat{\Sigma}$ is sample covariance of the surrogate's prediction error. This non-conformity score has a known analytical sublevel-set
\begin{equation*}
  \mathbb{C}^{\alpha}  = \{y \in \mathbb{R}^{n} \mid  s(x,y) \leq q(\alpha)\},
\end{equation*}
which is an ellipsoid centred at $\hat{f}(x)$, and eccentricity related to $\hat{\Sigma}$. They further show their method extends to `normalised' conformal prediction, where the ellipsoid changes depending on input $x$.

\subsection{Contributions}

Our contribution is most similar to the last of the above methods, where we predict a multivariate set equipped with a guaranteed $\alpha$-level frequentist performance. Our method diverges as we do not rely on a non-conformity score; instead, we directly construct prediction sets directly on a processed calibration data set. We summarise our contributions:

\begin{itemize}
  \item A conformal prediction method based on zonotopes
  % \item Methods to calibrate this representation from data using conformal prediction,
  \item An application of this technique to functional surrogate models, giving multivariate prediction sets for functional data,
  \item A method to account for the dimension reduction truncation error, ensuring the guarantees.
\end{itemize}

\section{Conformal prediction and consonant belief functions}\label{sec:conformal_prediction}
In this section we briefly describe inductive conformal prediction, for calibrating guaranteed prediction sets, particularly used for ML models (but not exclusively), and the related idea of belief functions, which we find to be a useful theory for performing computation using the calibrated sets.

\paragraph{Inductive conformal prediction} (ICP) is a computationally efficient version of conformal prediction \citep{papadopoulos2002inductive} for computing a set of possible predictions $\mathbb{C}^{\alpha}: X \mapsto \{\text{subsets of} \; Y\}$ of an underlying machine learning model $\hat{f}: X \mapsto Y$. The prediction set is equipped with the following probabilistic inequality
\begin{equation}\label{eq:conformal}
  \mathbb{P}(Y_{n+1} \in \mathbb{C}^{\alpha}(X_{n+1})) \geq 1 - \alpha.
\end{equation}
That is, the probability that the next unobserved prediction $Y_{n+1}$ is in computed set $\mathbb{C}^{\alpha}(X_{n+1})$ is bounded by $1-\alpha$, where $\alpha \in [0,1]$ is a user-defined error rate. Additionally, Equation~\ref{eq:conformal} can be \textit{guaranteed} if the data is \textit{exchangeable}, i.e. the data used to build $\mathbb{C}^{\alpha}$ can be replaced with the future unobserved samples $Y_{n+1}$ without changing their underlying distributions.

Most relevant for this paper is how $\mathbb{C}^{\alpha}$ is constructed, and how inequality~\ref{eq:conformal} is guaranteed. In ICP, one compares the predictions of a pre-trained model $\hat{f}$ to a \textit{calibration} dataset $Z = (Z_1, Z_2, \dots, Z_n)$ where $Z_i = (X_i, Y_i)$, which is yet unseen by $\hat{f}$. A non-conformity score $s: X \times Y \mapsto \mathbb{R}$ is used to compare $\hat{f}$'s predictions to $Z_i$, with large values of $s$ indicating a large disagreement between the prediction and the ground truth. A common score used in regression is $s(x,y) = |\hat{f}(x) - y|$. When applied to the calibration data, this yields non-conformity scores $\alpha_i = s(X_i, Y_i)$ for $i = 1, \ldots, n$.

The key insight is that these scores form an empirical distribution, and for a new test point, we can compute a p-value based on this empirical distribution. For a candidate prediction $y$ at test input $x$, the p-value is defined as the proportion of calibration scores that are at least as large as the test score:
\begin{equation}\label{eq:pvalues}
  p(x, y) = \frac{1}{n} \sum^{n}_{i=1}\mathbb{I}[s(x, y) \leq \alpha_{i}],
\end{equation} 
where $\mathbb{I}$ is the indicator function. Under exchangeability, this p-value has the property that $\mathbb{P}(p(X_{n+1}, Y_{n+1}) \leq \epsilon) \leq \epsilon$ for any $\epsilon \in [0,1]$.

The prediction set is then constructed by collecting all candidate values $y$ whose p-values exceed the significance level $\alpha$:
$$\mathbb{C}^{\alpha}(x) = \{y \in Y \mid p(x, y) > \alpha\}.$$
Equivalently, this can be expressed using a quantile of the empirical distribution of non-conformity scores: taking $q = \alpha_{\ceil{(1-\alpha)(n+1)}}$ (the $(1-\alpha)$-quantile of the empirical distribution), we have $\mathbb{C}^{\alpha}(x) = \{y \in Y \mid s(x,y) \leq q\}$. Note that the prediction regions form a nested family of sets w.r.t $\alpha$: $\mathbb{C}^{\alpha_1} \supseteq \mathbb{C}^{\alpha_2}$ for $0 \leq \alpha_1 \leq \alpha_2 \leq 1$.

\paragraph{Our framework} Computing the level-set of a complicated $s(x,y)$ function may be challenging, so often we opt for quite simple non-conformity scores with known level-sets, e.g., the level-set of $|y - \hat{f}(x)|$ is simply $[\hat{f}(x) \pm q]$. A challenge which is exacerbated for multivariate problems, somewhat limiting application. We therefore take a different approach, and directly construct $\mathbb{C}^{\alpha}$ sets using a parametric nested family of sets from the calibration data, a method originally suggested by~\cite{GUPTA2022108496}. This allows us to design prediction sets tailored to our particular data, and well-suited for multivariate problems. Using an interpretation of $\mathbb{C}^{\alpha}$ as belief functions, we can perform additional computations (for example linear and non-linear transformations) on $\mathbb{C}^{\alpha}$, while still maintaining the guarantee~\ref{eq:conformal}. This framework however comes with its own challenges. In some sense we are doing the reverse of conformal prediction, where our challenge is not to compute the level-set $\mathbb{C}^{\alpha} = \{y \in Y \mid s(x,y) \leq q(\alpha)\}$, rather we begin with $\mathbb{C}^{\alpha}$ and need to find the membership values $\alpha_i = \sup \{\alpha \in [0,1] \; \mid \; Z_i \in \mathbb{C}^{\alpha}\}$ of the calibration data. Depending on what set-representation is used, computing the membership values can be a costly operation.

\paragraph{Belief functions}\cite{cella2022validity} make a useful connection between conformal prediction and belief functions \citep{shafer1976mathematical}, a generalisation of the Bayesian theory of probability where one can make set-valued probabilistic statements, such as inequality~\ref{eq:conformal} given by conformal prediction.

Belief functions, also called random-sets or Dempster-Shafer structures, are set-valued random variables whose statistical properties (such their cdf, moments, sample realisations, and probability measure) are set-valued. Belief functions form a bound on a collection of partially unknown random variables, often used in robust risk analysis. In particular, the conformal nested prediction sets $\mathbb{C}^{\alpha}$ can be related a consonant belief function~\citep{DUBOIS1990419}, and under this framework transformations $f:X \mapsto Y$ of the imprecise random variable are quite simply:
\begin{equation*}
  \mathbb{C}_Y^{\alpha} = f(\mathbb{C}_X^{\alpha}),
\end{equation*}
that is, for each $\alpha$-level, a single set-propagation of $\mathbb{C}_X^{\alpha}$ through $f$ is required to determine $\mathbb{C}_Y^{\alpha}$, maintaining the same $\alpha$-confidence level. This is comparatively simpler than other representations, which we use to transform fitted $\mathbb{C}_X^{\alpha}$ through computations (SVDs in particular).
% \begin{definition}[Belief function]
%   A belief function is ...
% \end{definition}
% \begin{definition}[Consonant belief function]
%   A consonant belief function is a belief function whose focal elements are nested. That is, order by $\subseteq$.
% \end{definition}
% In particular, complex function evaluations 
\section{Zonotope prediction sets}
Before proceeding to functional data, we must describe how we will construct our multivariate prediction sets $\mathbb{C}^{\alpha}$. We will use \textit{zonotopes}—a class of convex sets offering advantageous properties for high-dimensional settings, including closure under linear transformations and Minkowski sums. Zonotopes generalise intervals, boxes, hyper-rectangles, and all their rotations, and may be represented on the computer in a compact manner. An $n$-dimensional zonotope is completely characterised by a vector in $\mathbb{R}^{n}$ (centre), and a collection of $p \in \mathbb{N}$ vectors in $\mathbb{R}^{n}$ (generators).
\begin{definition}[Zonotope]
  Given a centre $c_{\mathcal{Z}} \in \mathbb{R}^{n}$ and $p \in \mathbb{N}$ generator vectors in a generator matrix $G_{\mathcal{Z}} = [g_{1}, \dots, g_{p}] \in \mathbb{R}^{n \times p}$, a zonotope is defined as
  \begin{equation*}
    \mathcal{Z} = \left\{x \in \mathbb{R}^n \Big| \; x = c_{\mathcal{Z}} + \sum^{p}_{i=1} \xi_ig_i, \;\; \xi_i \in [-1, 1] \right\}
  \end{equation*} %https://people.kth.se/~kallej/papers/learning_pmlr21alan.pdf
  We will use the shorthand notation $\mathcal{Z} = \langle c_{\mathcal{Z}}, G_{\mathcal{Z}}\rangle$.
\end{definition}
Zonotopes can equally be characterised as the image of the box $[-1, 1]^p$ by an affine transformation $T(X) = c_{\mathcal{Z}} + G_{\mathcal{Z}}X$. Some example of 2D zonotopes and their generators are shown below (centre is omitted as it corresponds to a translation).

\begin{figure}[h!]
  \centering
  \begin{subfigure}[b]{0.156\textwidth}
    \centering
    \includegraphics[width=\textwidth]{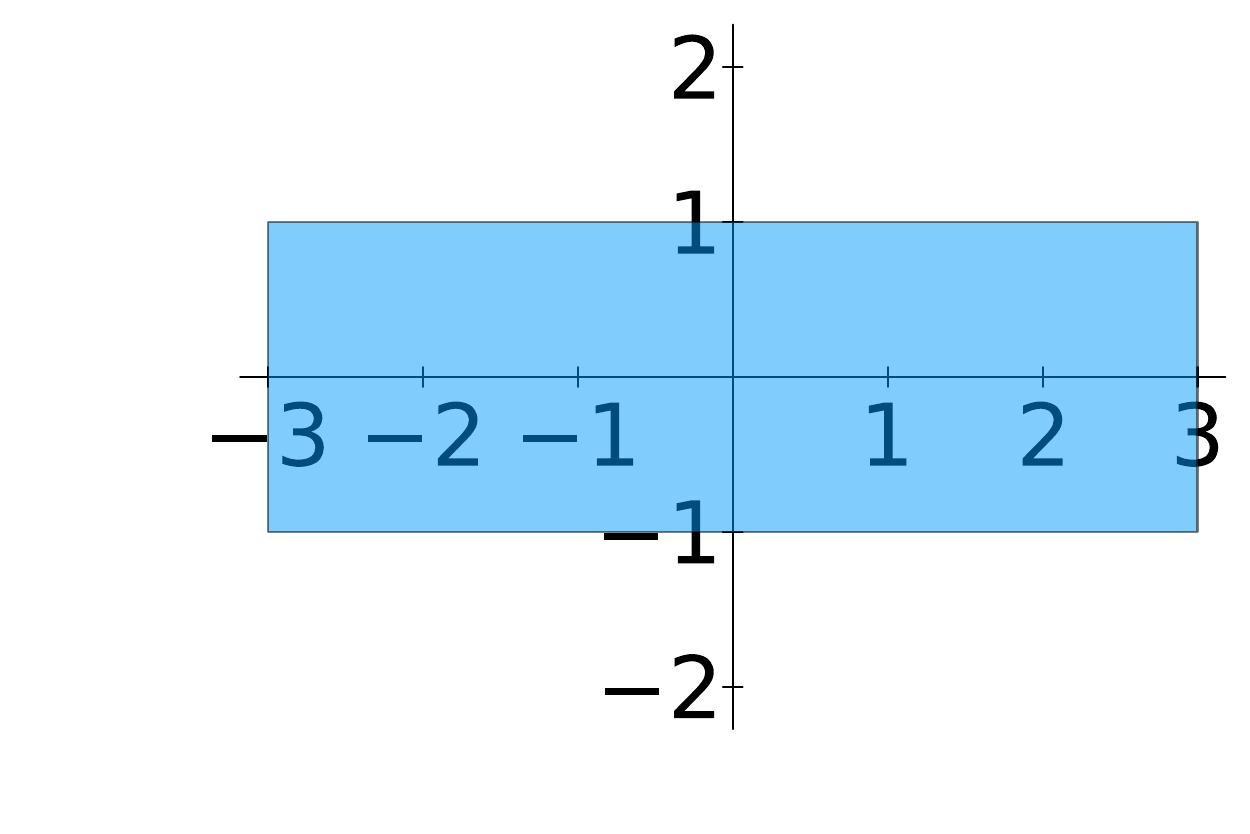}
    \[
      \begin{pmatrix}
      3 & 0 \\
      0 & 1
      \end{pmatrix}
      \]
\end{subfigure}
\begin{subfigure}[b]{0.156\textwidth}
  \centering
  \includegraphics[width=\textwidth]{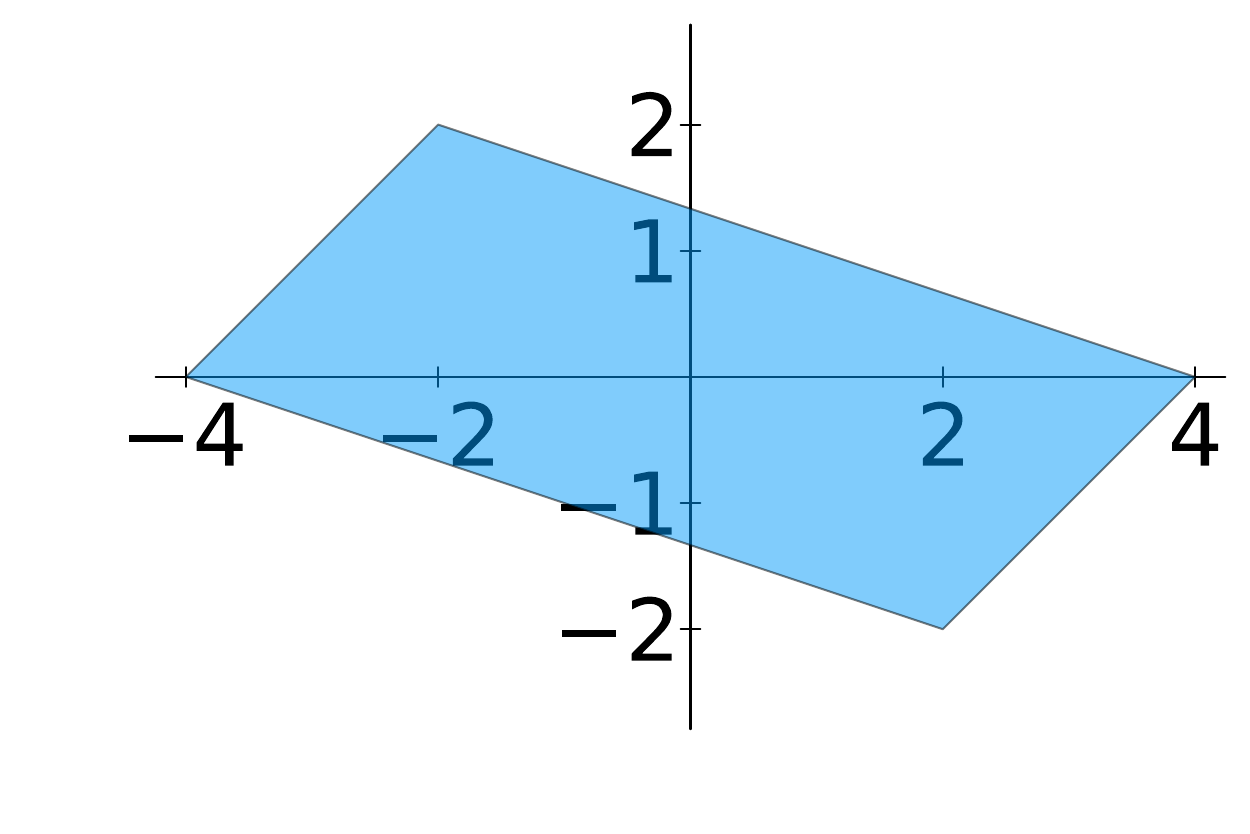}
  \[
    \begin{pmatrix}
    3 & 1 \\
    -1 & 1
    \end{pmatrix}
    \]
\end{subfigure}
\begin{subfigure}[b]{0.156\textwidth}
  \centering
  \includegraphics[width=\textwidth]{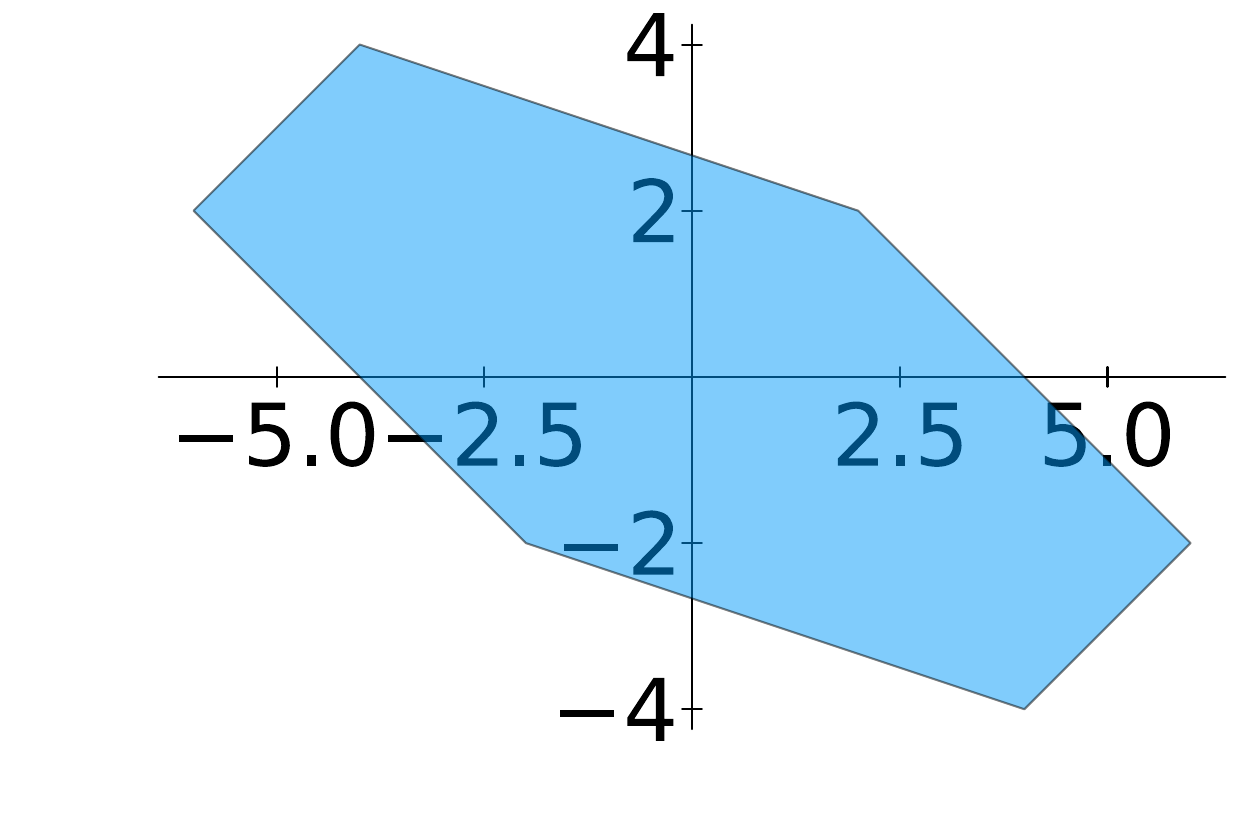}
  \[
    \begin{pmatrix}
    3 & 1 & -2\\
    -1 & 1 & 2
    \end{pmatrix}
    \]
\end{subfigure}
  \caption{Example zonotopes in blue, and their associated generator matrices immediately below them.}
  \label{fig:zonotope_examples}
\end{figure}
\begin{figure}[h!]
  \centering
  \begin{subfigure}[b]{0.156\textwidth}
    \centering
    \includegraphics[width=\textwidth]{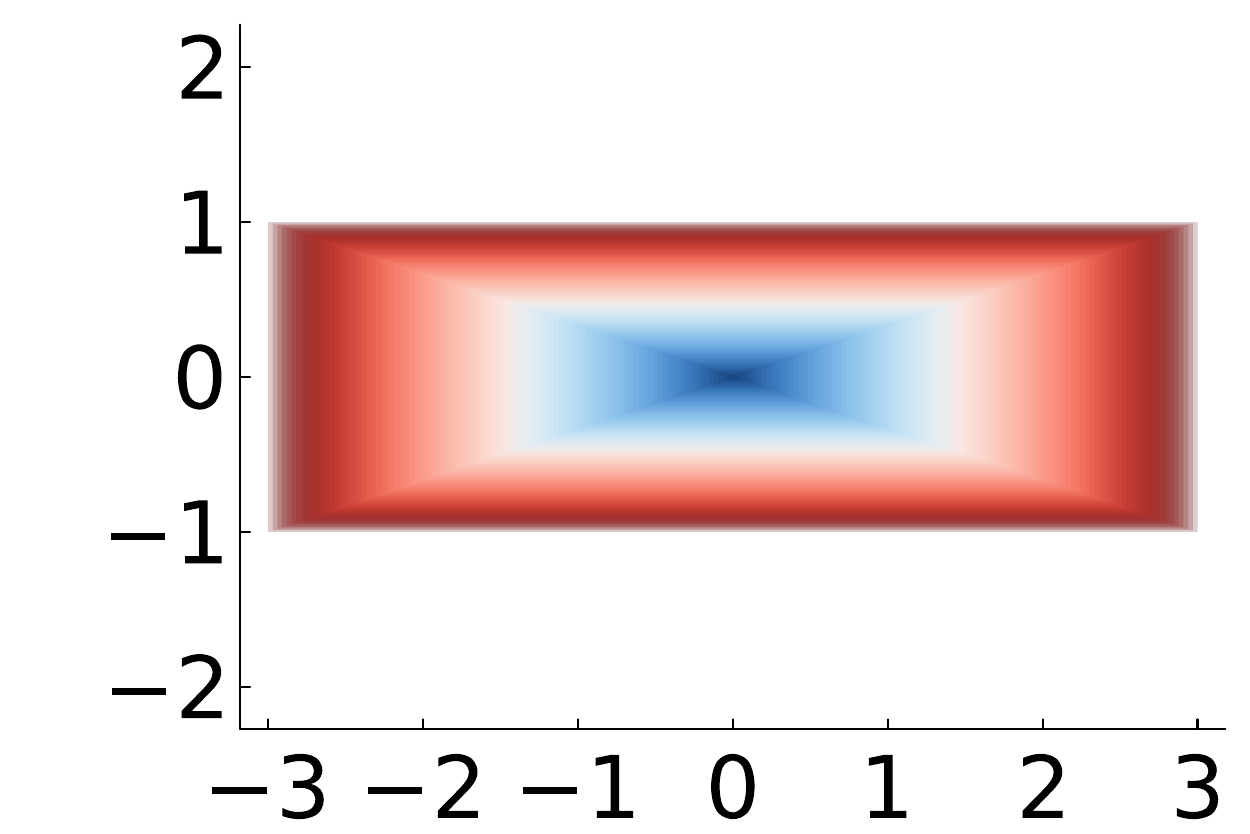}
      \[
        p_Z = \begin{pmatrix}
        0 \\
        0
        \end{pmatrix}
        \]
\end{subfigure}
\begin{subfigure}[b]{0.156\textwidth}
  \centering
  \includegraphics[width=\textwidth]{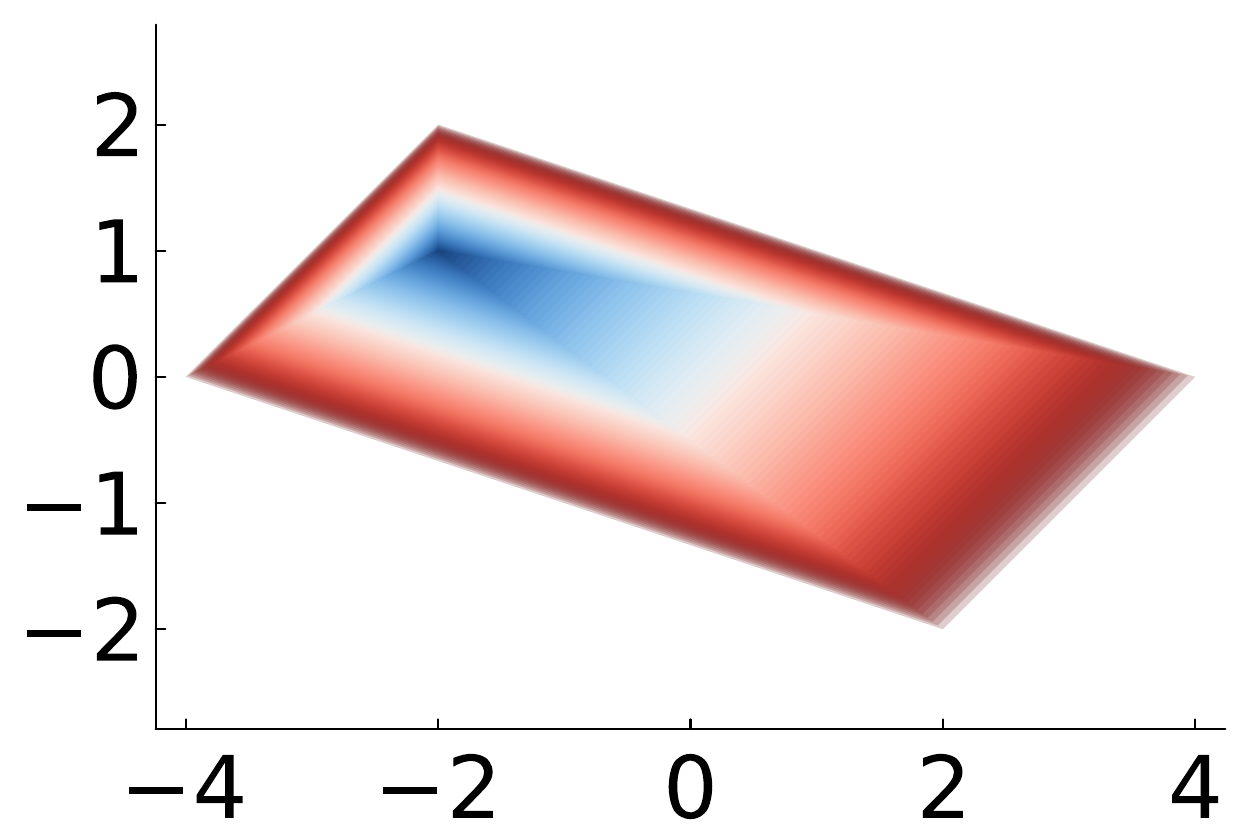}
    \[
    p_Z = \begin{pmatrix}
    -2  \\
    1
    \end{pmatrix}
    \]
\end{subfigure}
\begin{subfigure}[b]{0.156\textwidth}
  \centering
  \includegraphics[width=\textwidth]{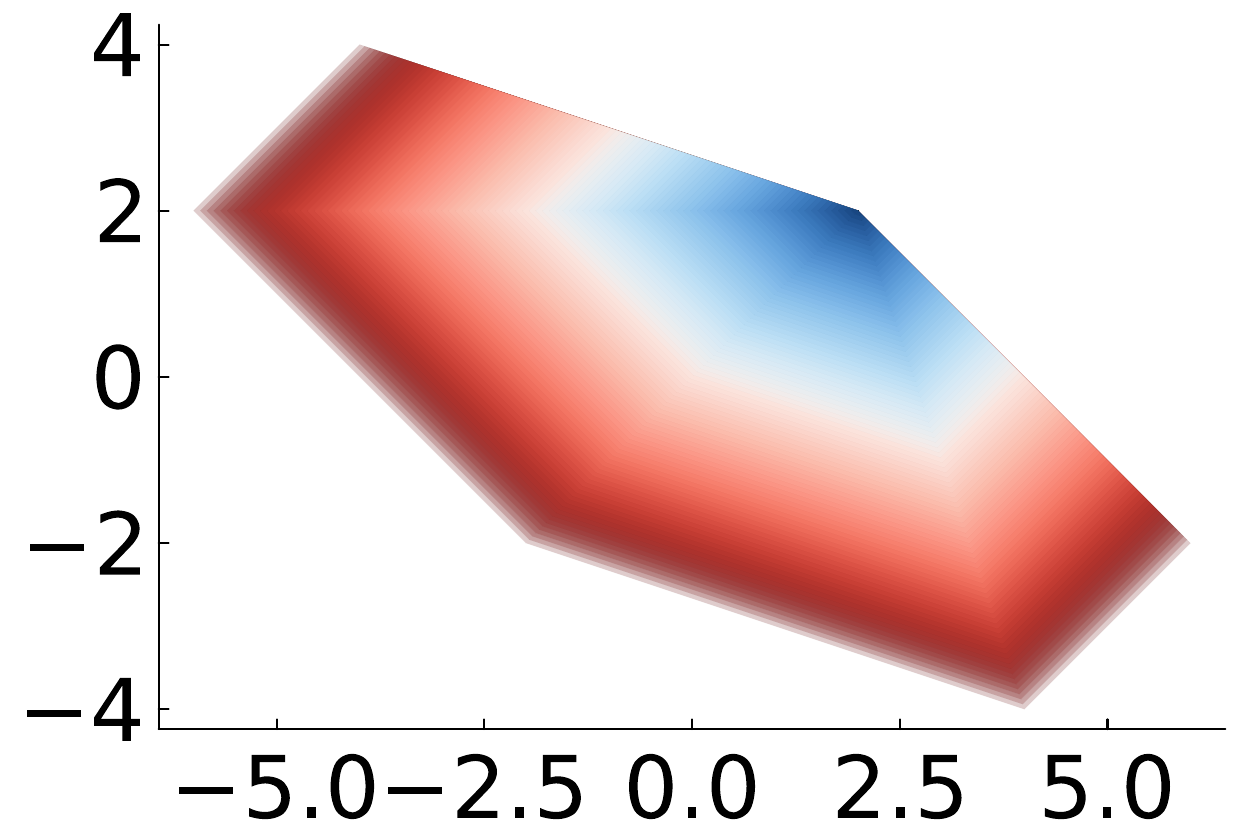}
    \[
      p_Z =\begin{pmatrix}
      2\\
      2 
      \end{pmatrix}
      \]
\end{subfigure}
  \caption{Examples of nested zonotopes families, with $\zonotope$ from Figure~\ref{fig:zonotope_examples} and indicated cores $p_Z$.}
  \label{fig:nested_topes}
\end{figure}

We will use zonotopes as a basis to construct nested prediction regions, giving the probabilistic guarantee ~\ref{eq:conformal}, calibrated with respect to a particular dataset. For this, we specify a parametric family of nested zonotopes, parameterised by a value $\alpha \in [0,1]$. We find that the following parameterisation is simple, and yields nested sets.
\begin{proposition}[nested zonotopic sets]\label{prop:nested_topes}
  Given a zonotope $\mathcal{Z} = \langle c_{\mathcal{Z}}, G_{\mathcal{Z}}\rangle$, a point $p_{z} \in \mathcal{Z}$, the following collection of zonotopes are nested 
  \begin{equation*}
      \mathcal{Z}^{\alpha}_{p_{Z}} = \langle \; c_{\mathcal{Z}}(1-\alpha) + p_{Z}\alpha, \; G_{\mathcal{Z}} (1- \alpha) \;\rangle,
  \end{equation*}
    with $\alpha \in [0, 1]$. $\mathcal{Z}^{\alpha}_{p_{Z}}$ are nested in the sense that $\mathcal{Z}^{\alpha_1}_{p_{Z}} \supseteq \mathcal{Z}^{\alpha_2}_{p_{Z}}$ for any $\alpha_1 \leq \alpha_2$.
\end{proposition}
A detailed proof that $\mathcal{Z}^{\alpha}_{p_{Z}}$ forms a nested collection of sets in found in appendix~\ref{app:proof}. \textbf{A quick sketch of the proof} is that we show that the half-spaces $H = \{x \in \mathbb{R}^p | a{\top}x \leq b\}$ composing the box $[-1,1]^p$ are nested $H^{\alpha_1} \supseteq H^{\alpha_2}$ for $\alpha_1 \leq \alpha_2$ under the transformation $c_{\mathcal{Z}}(1-\alpha) + p_{Z}\alpha + G_{\mathcal{Z}}(1- \alpha) X$. We also note that $\mathcal{Z}^{\alpha=0}_{p_{Z}} = \mathcal{Z}$, (the largest set in the family), and $\mathcal{Z}^{\alpha=1}_{p_{Z}} = \{p_{Z}\}$  Several examples of parametric nested zonotopes are shown in Figure~\ref{fig:nested_topes}.

\paragraph{Why zonotopes} Although Proposition~\ref{prop:nested_topes} is simple (and its proof not technically deep), the choice of zonotopes is motivated by two key properties:
%that make them scalable to our setting:
\begin{itemize}
\item \textbf{Closure under Cartesian products.} The Cartesian product of zonotopes (or with boxes) yields another zonotope. We use this property in step 5 above (section~\ref{sec:steps}) when bounding the truncation error of an SVD.
\item \textbf{Efficiency under linear transformations.} Zonotopes can be projected exactly through linear maps via a matrix multiplication. This makes step 6 above very efficient, where we map uncertainty sets back to the prediction space of the surrogate.
\end{itemize}
Together, these properties make zonotopes a scalable and tractable choice for constructing prediction sets in high-dimensional and function-valued cases.

\begin{figure}[h!]
  \centering
  \begin{subfigure}[b]{0.156\textwidth}
    \centering
    \[
      \mathcal{Z}^{s(\alpha)}_{p_Z}
      \]
    \includegraphics[width=\textwidth]{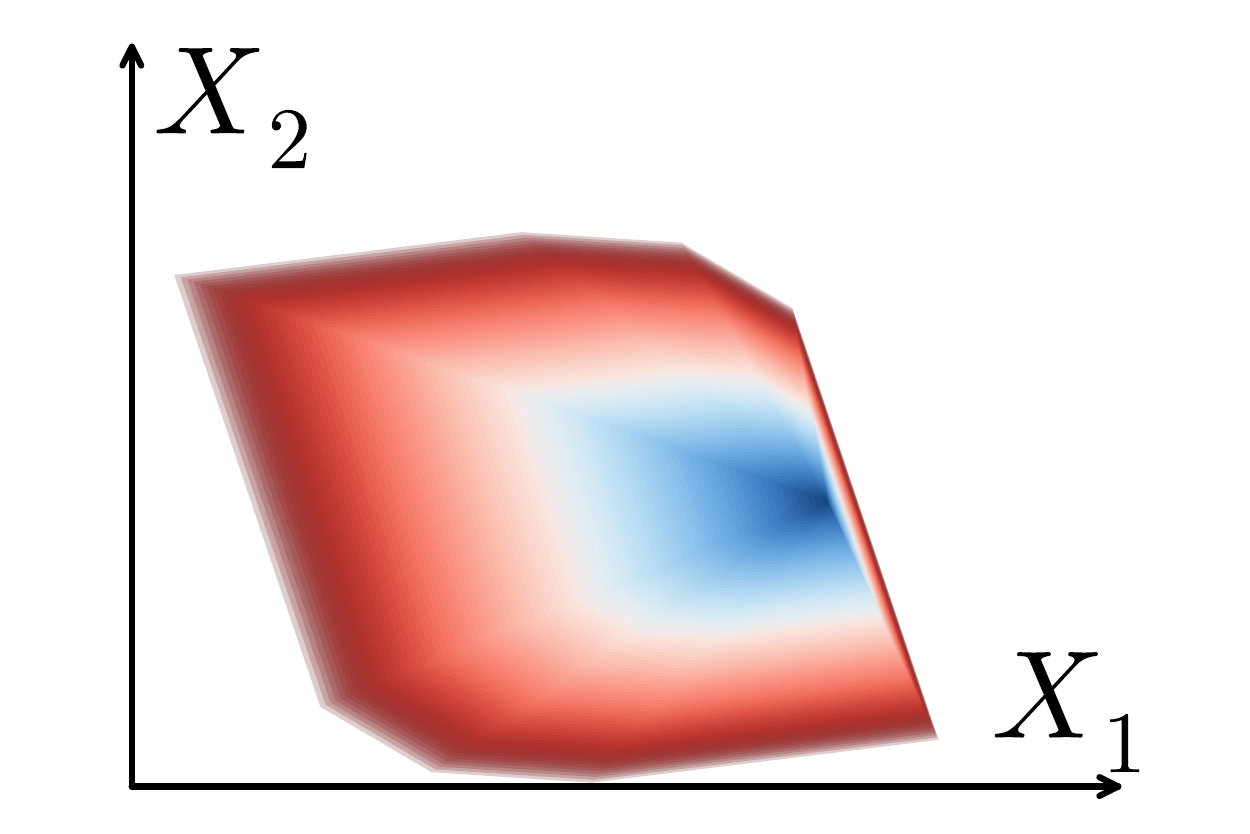}
\end{subfigure}
\begin{subfigure}[b]{0.156\textwidth}
  \centering
  \[
  \text{marginal } X_1
  \]
  \includegraphics[width=\textwidth]{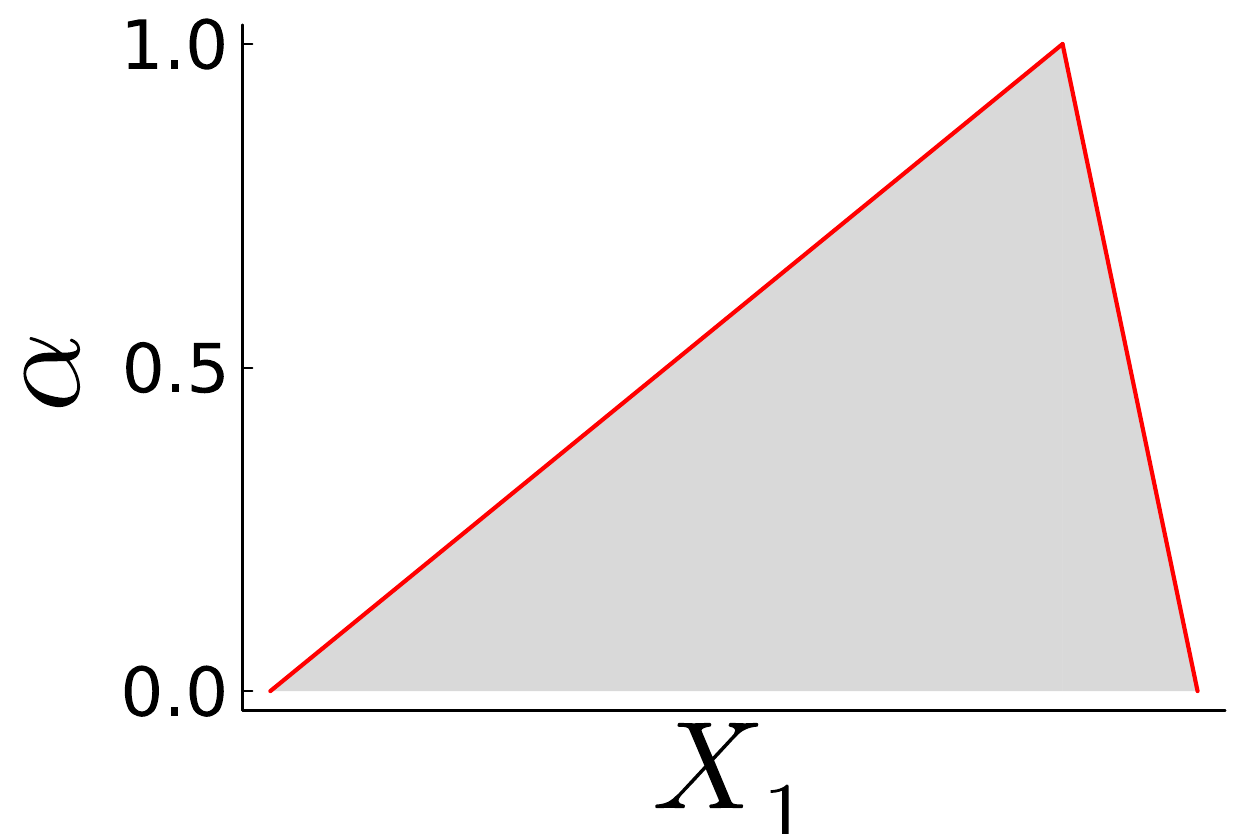}
\end{subfigure}
\begin{subfigure}[b]{0.156\textwidth}
  \centering
  \[
      s(\alpha)
      \]
  \includegraphics[width=\textwidth]{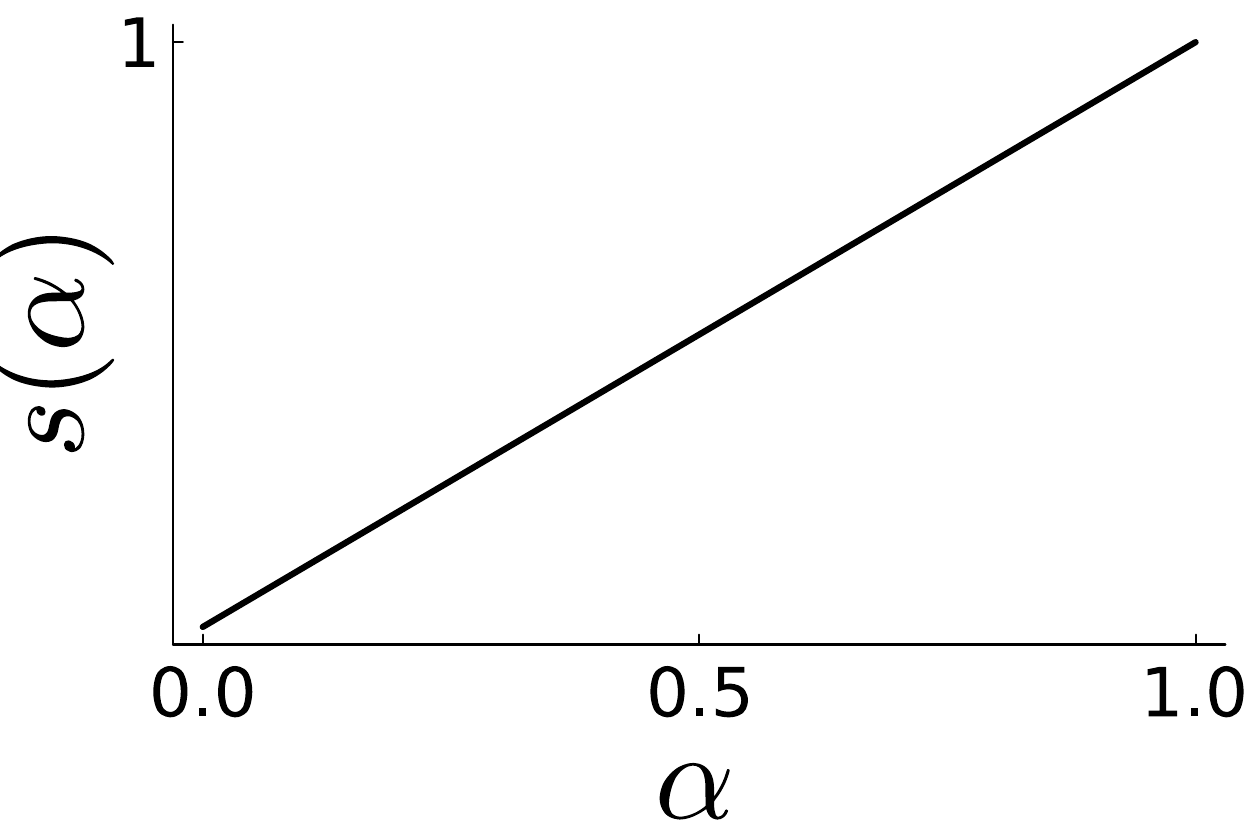}
\end{subfigure}

\begin{subfigure}[b]{0.156\textwidth}
  \centering
  \includegraphics[width=\textwidth]{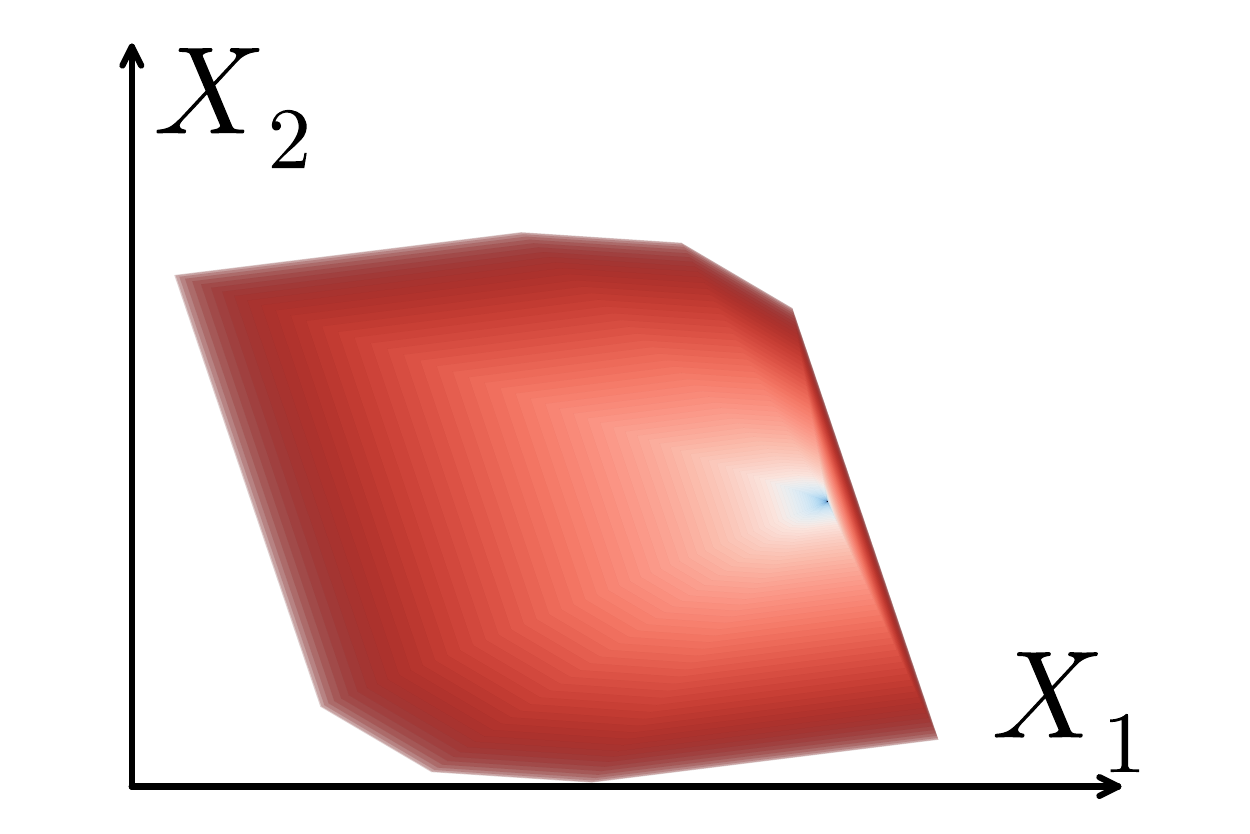}
\end{subfigure}
\begin{subfigure}[b]{0.156\textwidth}
\centering
\includegraphics[width=\textwidth]{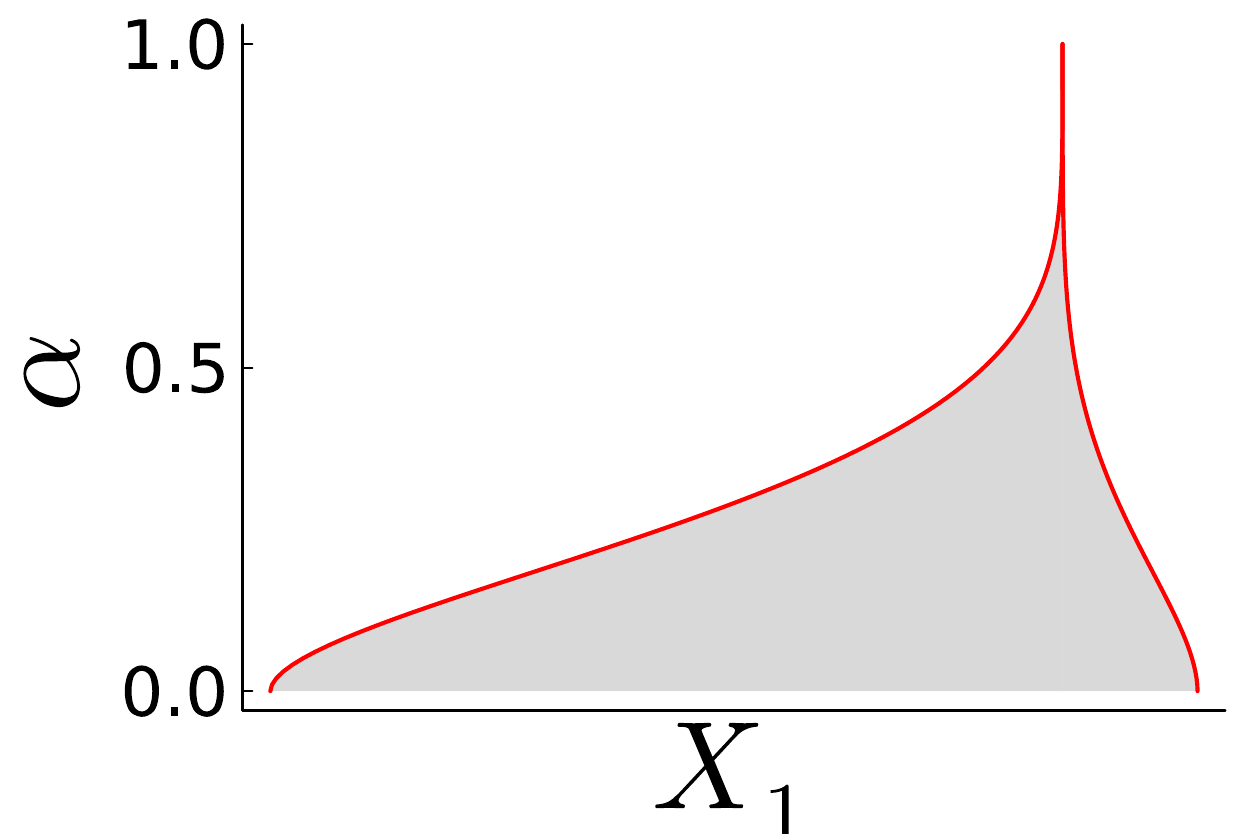}
\end{subfigure}
\begin{subfigure}[b]{0.156\textwidth}
\centering
\includegraphics[width=\textwidth]{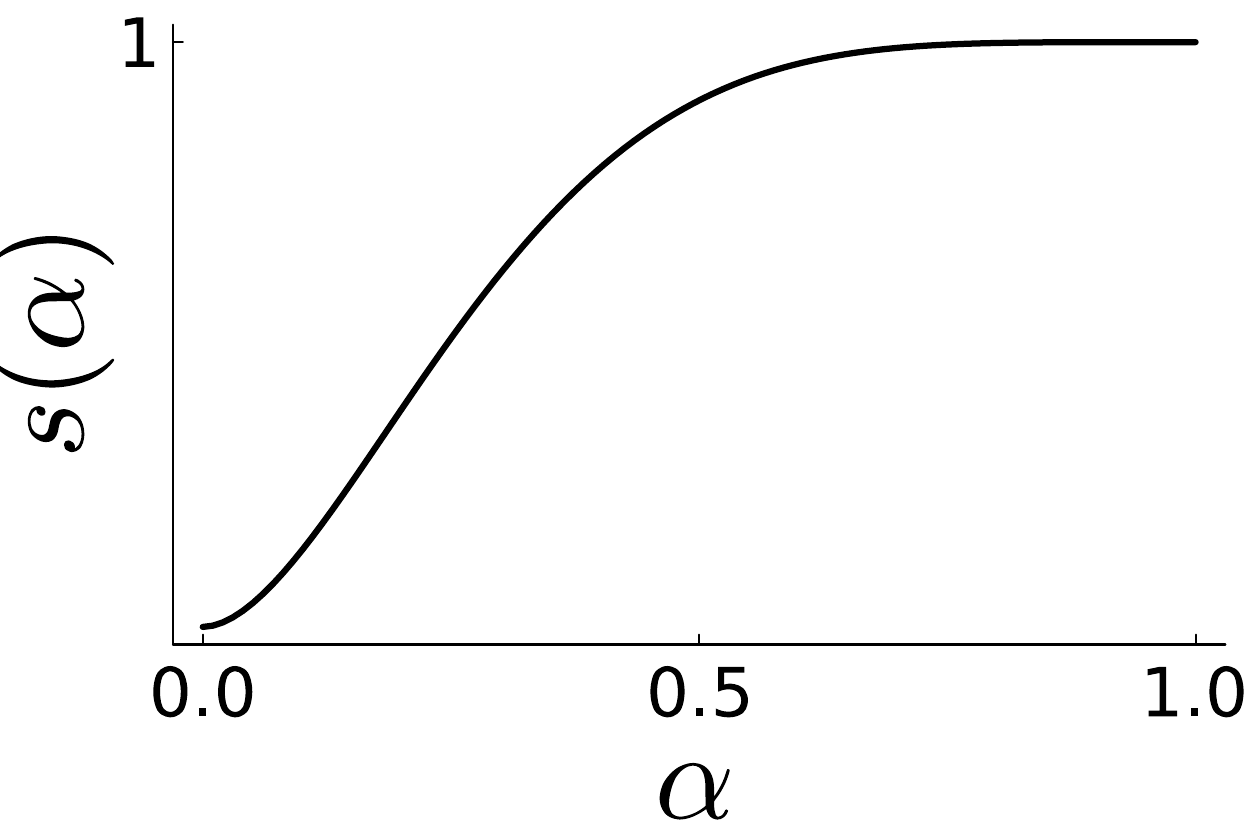}
\end{subfigure}
\begin{subfigure}[b]{0.156\textwidth}
  \centering
  \includegraphics[width=\textwidth]{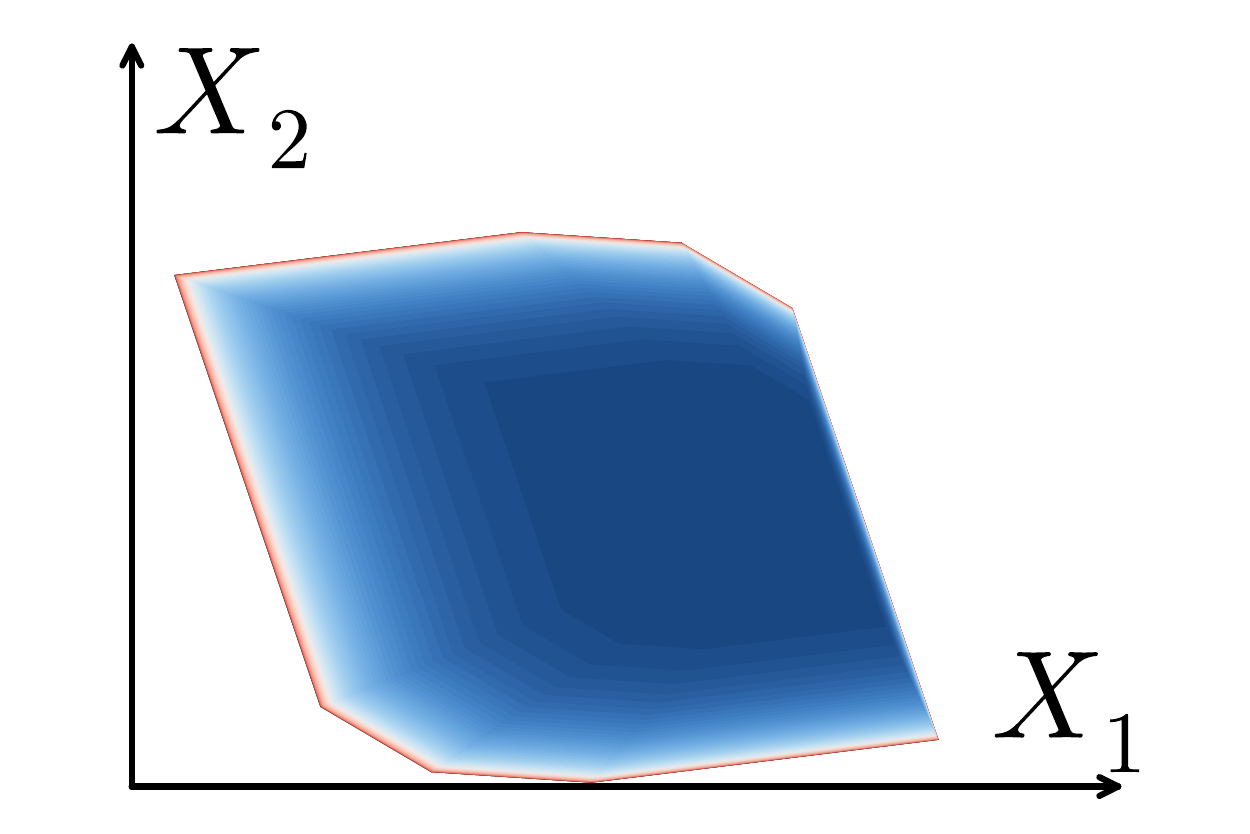}
\end{subfigure}
\begin{subfigure}[b]{0.156\textwidth}
\centering
\includegraphics[width=\textwidth]{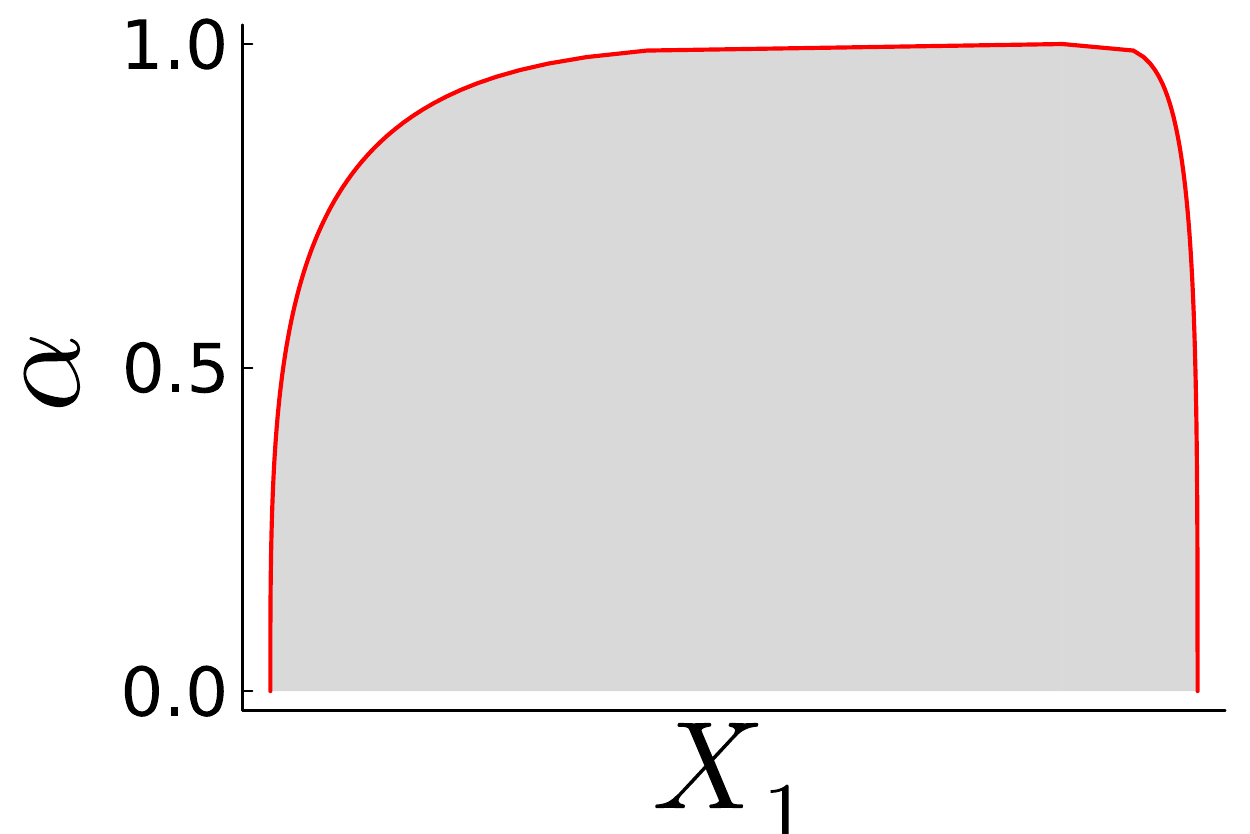}
\end{subfigure}
\begin{subfigure}[b]{0.156\textwidth}
\centering
\includegraphics[width=\textwidth]{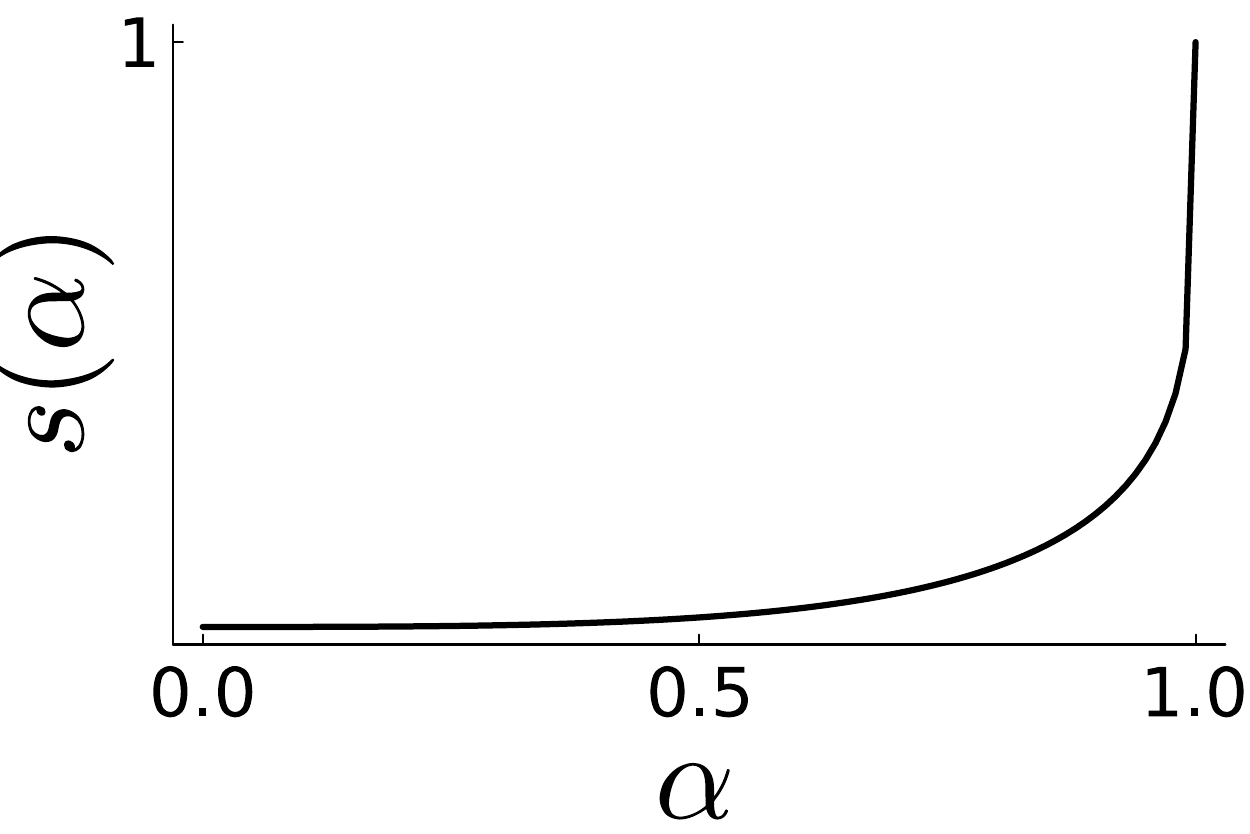}
\end{subfigure}
  \caption{Three examples of the same nested zonotope family $\nestedzonotope$, but with different $s(\alpha)$ functions. The central column shows the projection of the sets onto the $X_1$ dimension.}
  \label{fig:different_alphas}
\end{figure}

Note that the zonotopes $\nestedzonotope$ are not yet calibrated to anything, so the desired property $\mathbb{P}(X \in \nestedzonotope) \geq 1 - \alpha$ does not yet hold. The $\alpha$ is so far just a parameter, and needs to be related to a random variable $X$ of interest. In Section~\ref{sec:calibration} we explore a method to probabilistically calibrate $\nestedzonotope$ using conformal prediction. This essentially boils down to finding a monotonic function $s: [0, 1] \mapsto [0, 1]$ for $\alpha$ such that the inequality~\ref{eq:conformal} is guaranteed. Indeed, the $\nestedzonotope$ shown in Figure~\ref{fig:nested_topes} show a uniformly evaluated $\alpha$. If one replaces this with a monotonic function $s(\alpha)$,  quite dramatically different nested structures are obtained. Examples are shown in Figure~\ref{fig:different_alphas}. One may interpret $s(\alpha)$ as the cumulative probability distribution (cdf) of $\alpha$, which when sampled yields a random zonotope.

We note that inequality~\ref{eq:conformal} can be guaranteed irrespective of the chosen $\zonotope$ and $p_Z$, however the `quality' of the prediction sets (their size relative to their $\alpha$ confidence-level) depends on how well $\nestedzonotope$ captures $X$'s distribution shape. We therefore require the \textit{shape} of the $\nestedzonotope$ sets be \textit{fitted} with respect to the dataset. That is given, some data $\{X_1,\dots,X_N\}$ sampled from an unknown distribution $X_i \sim F_{X}$, we would like to find an enclosing zonotope such that all $X_i \in \zonotope$. We also require the core $p_Z$ to be fitted, which plays the role of defining the region of the highest confidence, the `central' point of the dataset in some sense. That is, all the prediction sets will contract towards this point, and so it is desirable for the region around this point to occupy a high density of samples. This point can be determined in terms of \textit{data depth}, how deep a point is in a dataset with respect to some metric. This point is somewhat analogous to the Bayesian posterior mode. Methods to fit $\nestedzonotope$ are described in Section~\ref{sec:fitting}.

A potentially interesting corollary of proposition~\ref{prop:nested_topes} are a simpler family of nested hyperrectangles $\mathcal{B}^{\alpha}_{p_B}$, in terms of a centre vector $c_{\mathcal{B}} \in \mathbb{R}^{n}$, radius vector $r_{\mathcal{B}} \in \mathbb{R}^{n}$, core $p_{B} \in \mathcal{B}$, and $\alpha \in [0,1]$.
\begin{corollary}
  The following family of hyperrectangles are nested
  \begin{equation*}
      \mathcal{B}^{\alpha}_{p}= \left\{x \in \mathbb{R}^n \Big| \; |x_i - (1-\alpha)c_i - \alpha p_i| \leq (1-\alpha)r_i  \right\},
  \end{equation*}
  for all $i=1,\dots,n$, and with $p \in \mathcal{B}$ and $\alpha \in [0, 1]$.
\end{corollary}
Some analysis is simpler in terms of hyperrectangles, and so one could opt to use this family rather than $\nestedzonotope$.

\subsection{Fitting zonotope prediction sets}\label{sec:fitting}
In this section we discuss various methods to fit $\nestedzonotope$ to a dataset $X_i \sim F_{X}$, that is to find a data-enclosing zonotope $X_i \in \zonotope$ for all samples, and an estimated point $p_Z$ with a high data-depth.

\paragraph{Fitting a rotated hyperrectangle for $\zonotope$} A simple but effective method is to enclose $X_i$ 
is using rotated hyperrectangle, with generators $G_{\mathcal{Z}}$ along the principal components of the data.

% Use figure* to span both columns
\begin{figure}[h!]
  \centering
  % First row (plots 1 and 2)
  \begin{minipage}{0.2\textwidth}
      \centering
      \includegraphics[trim={2.5cm 2.5cm 2.5cm 2.5cm}, width=\linewidth]{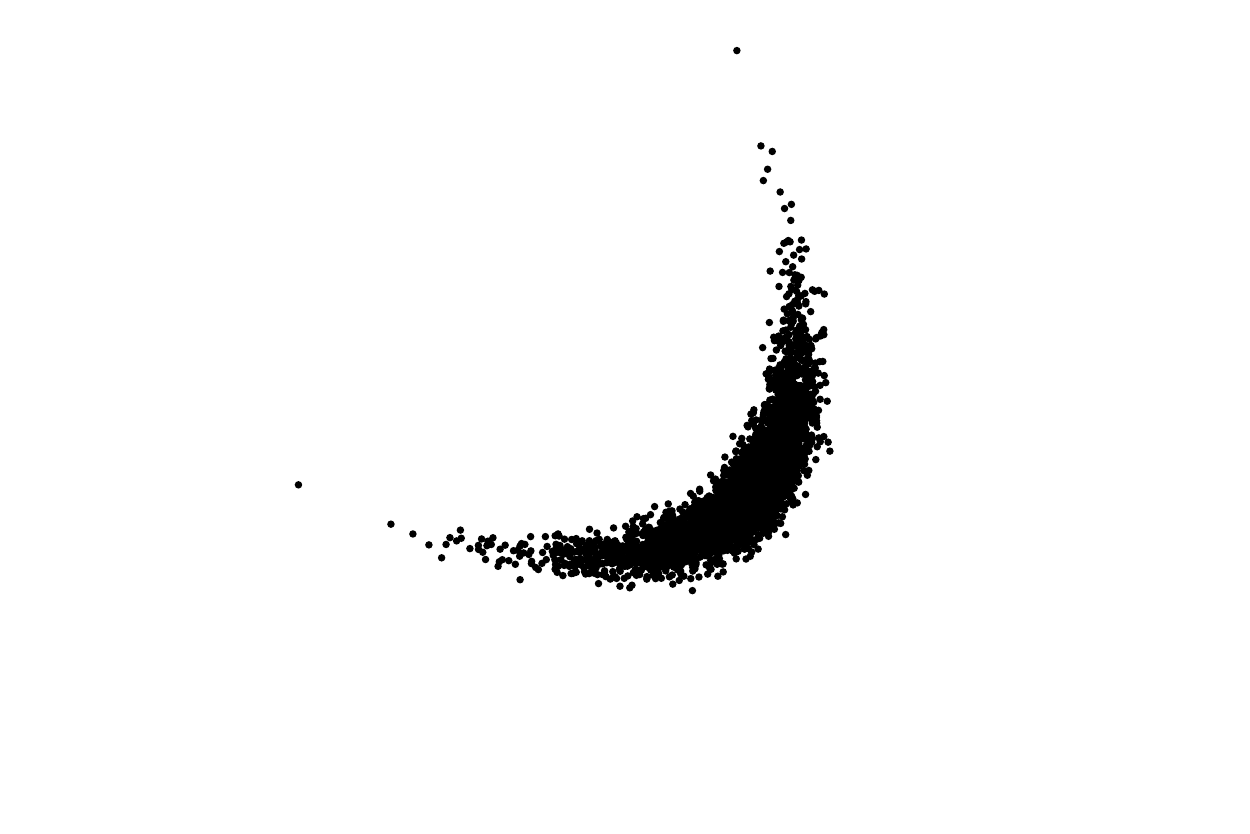} % Plot 1
  \end{minipage}%
  \hfill
  \begin{minipage}{0.2\textwidth}
      \centering
      \includegraphics[width=\linewidth]{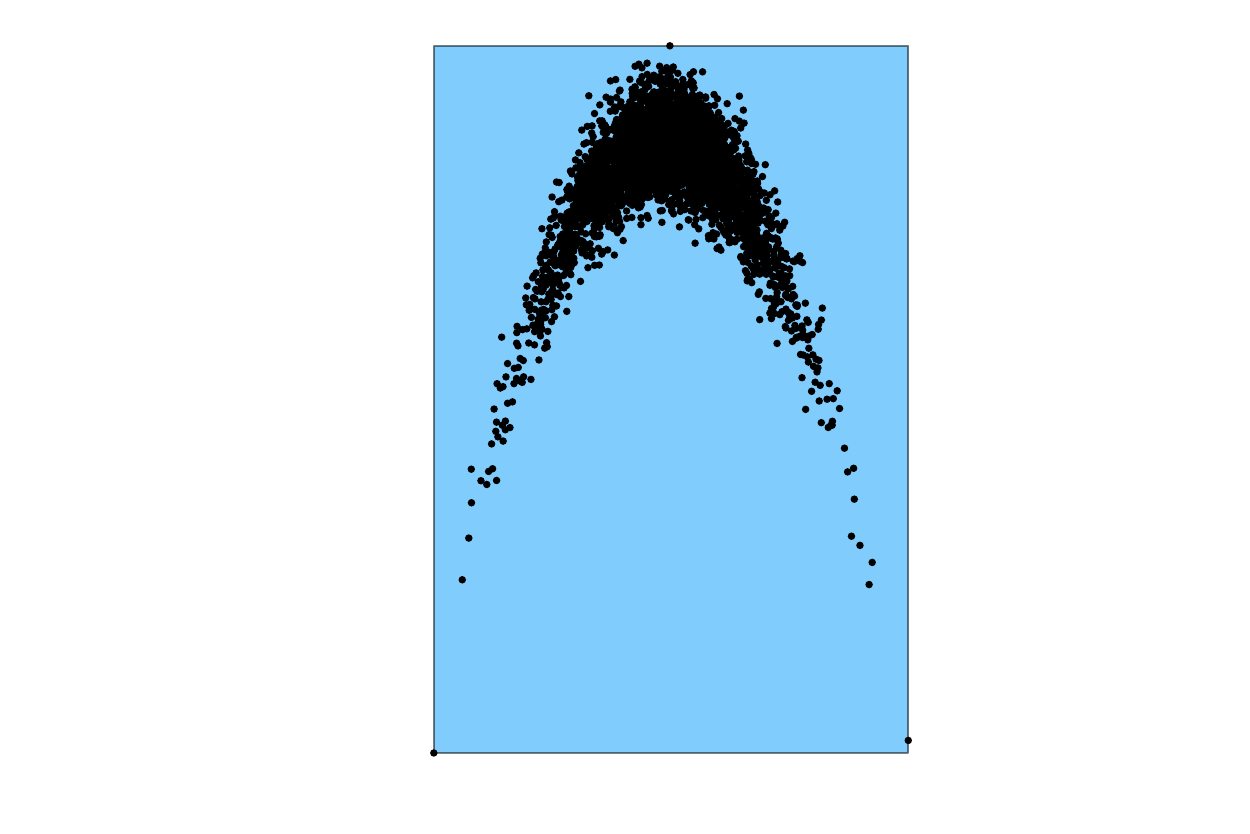} % Plot 2
  \end{minipage}

  \vspace{1.6cm} % Vertical space between rows

  % Second row (plots 3 and 4)
  \begin{minipage}{0.2\textwidth}
    \centering
    \includegraphics[trim={2.5cm 2.5cm 2.5cm 2.5cm}, width=\linewidth]{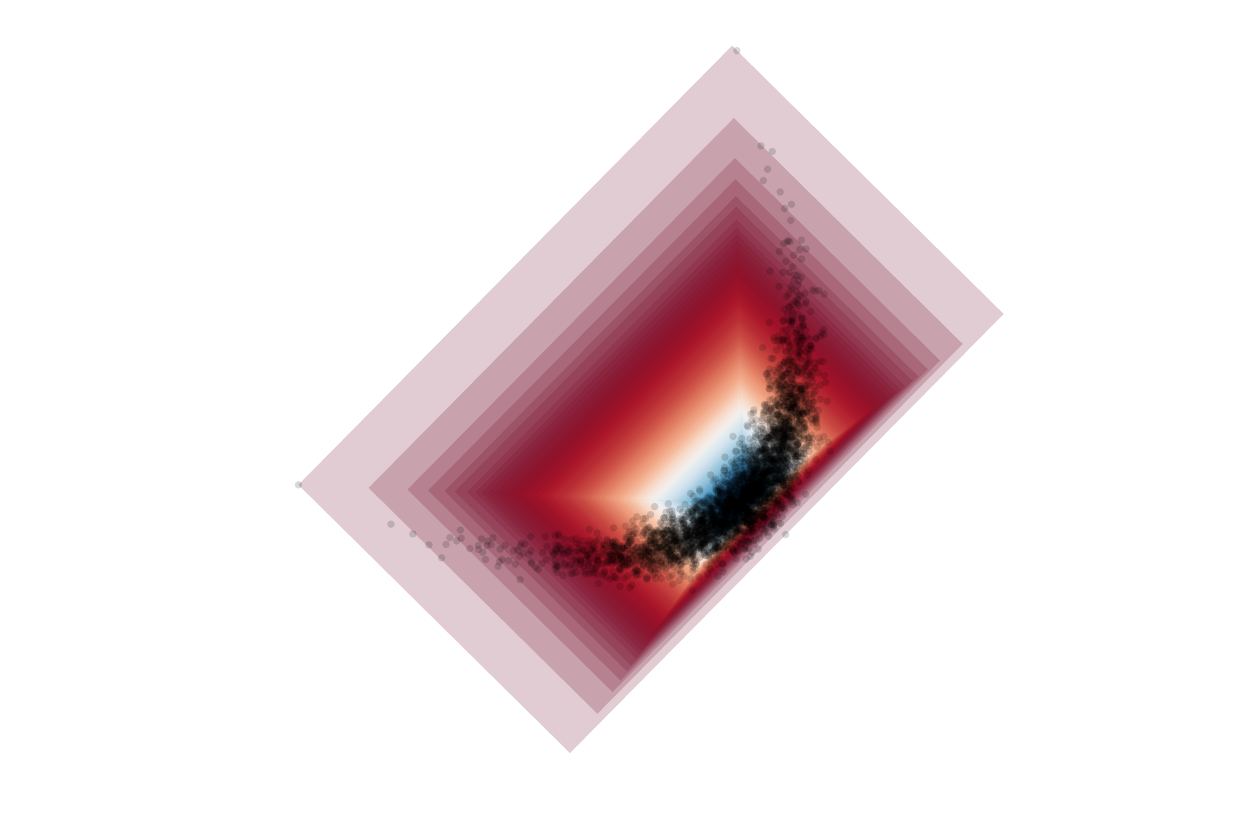} % Plot 4
  \end{minipage}  
  \hfill
  \begin{minipage}{0.2\textwidth}
    \centering
    \includegraphics[trim={2.5cm 2.5cm 2.5cm 2.5cm}, width=\linewidth]{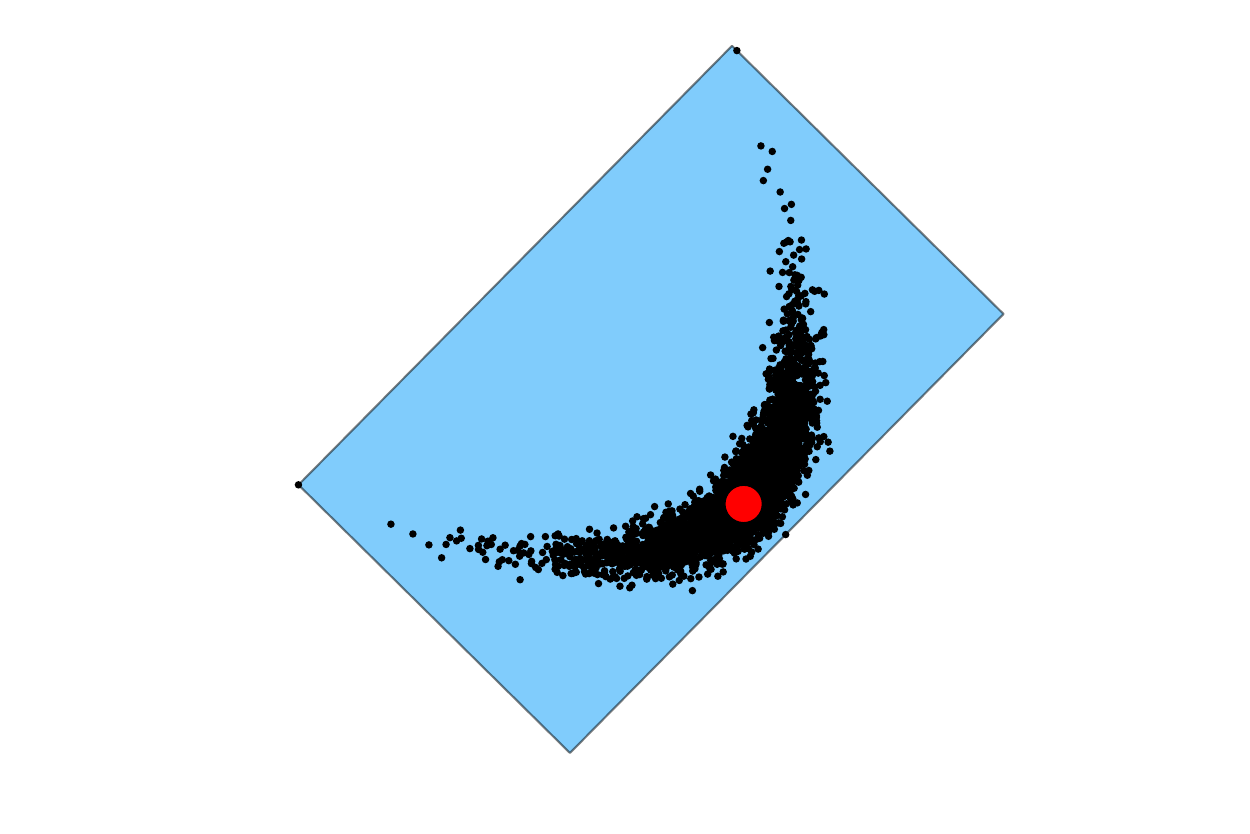} % 
  \end{minipage}%

  % Adding arrows with updated coordinates
  \begin{tikzpicture}[overlay, remember picture]
    % Arrow from Plot 1 (top-left) to Plot 2 (top-right)
    \draw[->, line width=0.5mm] (-1, 4) -- (1, 4) % Arrow from Plot 1 to Plot 2
    node[midway, above, align=center] {SVD \\ \& \\ bounding box}; % Add annotation at the midpoint
    % Arrow from Plot 2 (top-right) to Plot 3 (bottom-right)
    \draw[->, line width=0.5mm] (2.5, 3.5) -- (2.5, 2.5) % Arrow from Plot 2 to Plot 3
    node[midway, right, align=center] {Project \\ \& \\ find $p_Z$}; % Add annotation to the right of the arrow
    % Arrow from Plot 3 (bottom-right) to Plot 4 (bottom-left)
    \draw[->, line width=0.5mm] (1, 1.25) -- (-1, 1.25) % Arrow from Plot 3 to Plot 4
    node[midway, below, align=center] {Calibrate}; % Add annotation below the arrow
  \end{tikzpicture}
  \caption{Illustration of fitting a zonotope using the principal component analysis. The data is rotated into the PCA basis, a bounding hyperrectangle is fit in this space, then mapped back to the original coordinates. The resulting zonotope has generators aligned with the principal components, capturing the dominant directions of variation.}
  \label{fig:bounding_box}
\end{figure}
A singular value decomposition (SVD) of the data $X = U\Sigma V^{\top}$ yields a diagonal matrix $\Sigma$ of (decreasing) singular values and $V^{\top}$, the matrix of right singular vectors. The singular values provide a natural ranking of each eigenvector's contribution to the variance of $X$. An enclosing hyperrectangle can be easily found by computing the data $\min$ and $\max$ over each dimension of $U$. Upon converting this hyperrectangle (more details appendix~\ref{app:hyperrect}) to a zonotope representation, the zonotope $\zonotope_U = \langle c_{U}, G_{U}\rangle$ can be projected back to the original space by $\zonotope_X =  \langle V \Sigma c_{U}, V \Sigma G_{U}\rangle$. This yields a zonotope whose generators are aligned with the eigenvectors of the dataset. Figure~\ref{fig:bounding_box} illustrates this on a half-moon dataset. Although simple, quick to compute, and scalable, an obvious downside to this method is that it only yields zonotopes with generators as the same number of dimensions of $X$.

\begin{figure}[h!]
  \centering
  \begin{tikzpicture}
    % First image
    % \node (img1) at (0,0) {\includegraphics[trim={2.2cm 2.2cm 2.2cm 2.2cm}, width=0.156\textwidth]{figures/Guassian_1.pdf}};
    
    % Second image
    \node (img2) at (0,0) {\includegraphics[trim={2.2cm 2.2cm 2.2cm 2.2cm}, width=0.19\textwidth]{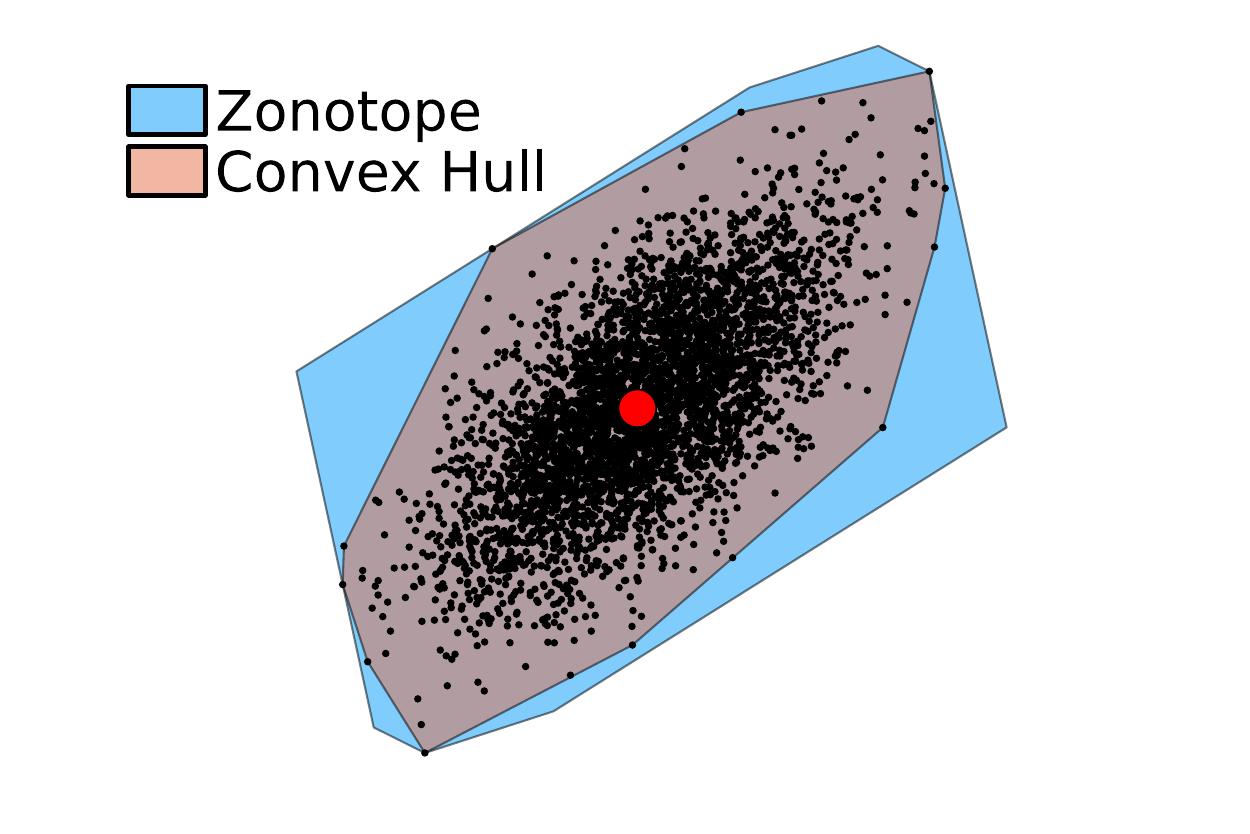}};
    
    % Third image
    \node (img3) [right=of img2, xshift=0.3cm] {\includegraphics[trim={2.2cm 2.2cm 2.2cm 2.2cm}, width=0.19\textwidth]{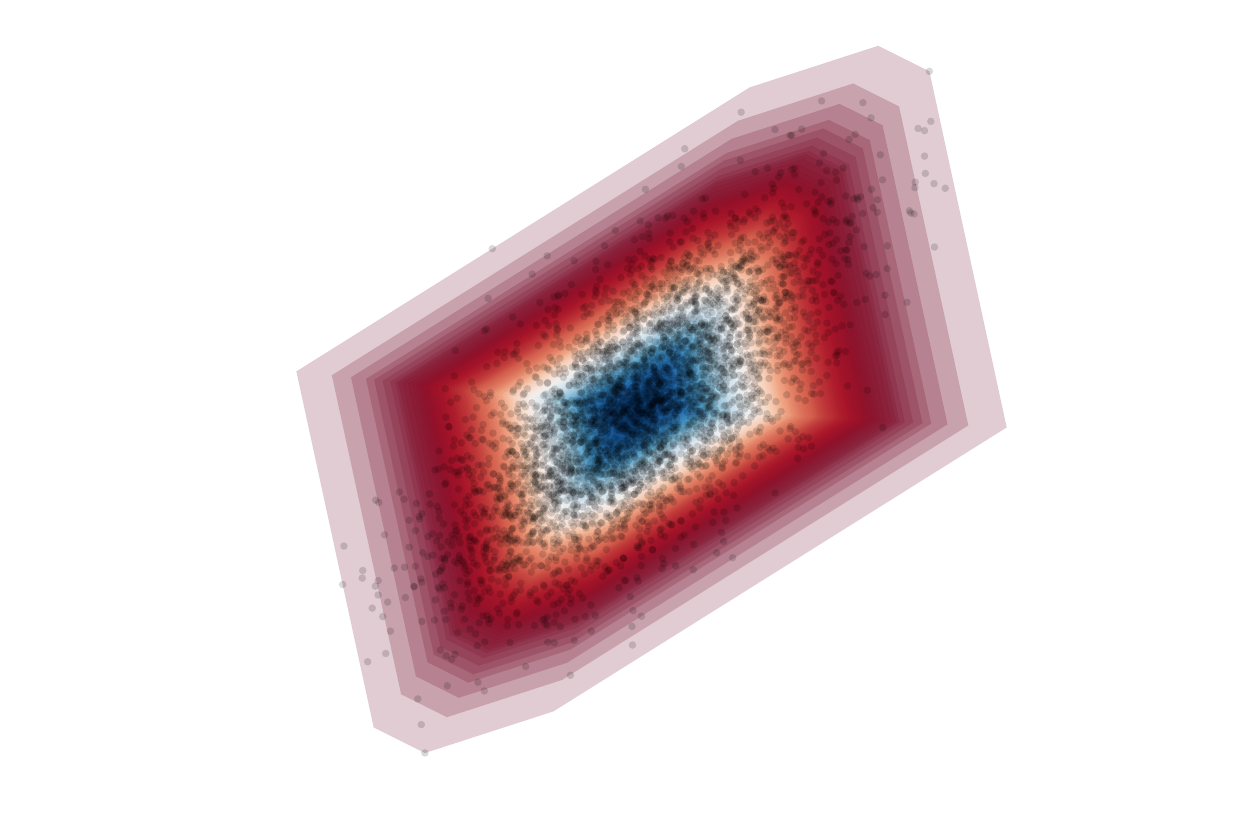}};
    
    % Arrows
    % \draw[->, thick] (img1.east) -- (img2.west) 
    %   node[midway, above, align=center] {Convex hull}
    %   node[midway, below, align=center] {$\zonotope$ \& $p_z$};

    \draw[->, thick] (img2.east) -- (img3.west) 
      node[midway, above] {Calibrate};
  \end{tikzpicture}
  \caption{Illustration of fitting using the convex hull.}
  \label{fig:bounding_convex_hull}
\end{figure}

\paragraph{Overapproximating a convex hull for $\zonotope$} A second method to fit $\zonotope$ is by first computing the convex hull $C_H$ of $X_i$, giving a bounding polytope of the data. An enclosing zonotope $\zonotope \supseteq C_H$ can then be computed (detailed in appendix~\ref{app:poly}). This yields a zonotope with half the number of generators as there are bounding half-spaces of $C_H$. We note that the overapproximation of a convex hull with a zonotope requires a conversion between the vertex representation (v-rep) of polytope to the half-space representation (h-rep), where the v-rep corresponds to a vector of extreme points, and the h-rep is a vector of bounding half-spaces. The conversion from v-rep to h-rep (and vice-versa) can be an expensive operation. Figure~\ref{fig:bounding_convex_hull} illustrates this method on a correlated Gaussian distribution.
% \begin{figure}[h!]
%   \centering
%   \begin{subfigure}[b]{0.156\textwidth}
%     \centering
%     \includegraphics[trim={2.2cm 2.2cm 2.2cm 2.2cm}, width=\textwidth]{figures/Guassian_1.pdf}
%   \end{subfigure}
%   \begin{subfigure}[b]{0.156\textwidth}
%     \centering
%     \includegraphics[trim={2.2cm 2.2cm 2.2cm 2.2cm}, width=\textwidth]{figures/Gaussian_zonotope.pdf}
%   \end{subfigure}
%   \begin{subfigure}[b]{0.156\textwidth}
%     \centering
%     \includegraphics[trim={2.2cm 2.2cm 2.2cm 2.2cm}, width=\textwidth]{figures/calibrated_zonotope_gaussian.pdf}
%   \end{subfigure}
%   % Adding manually positioned arrows using TikZ
%   \begin{tikzpicture}[overlay, remember picture]
%     % Arrow from Figure 1 to Figure 2 (manual positioning)
%     \draw[->, thick] 
%       (-3,0) -- (-1.5,0) node[midway, above, align=center] {Convex hull}
%       node[midway, below, align=center] {$\zonotope$ \& $p_z$};

%     % Arrow from Figure 2 to Figure 3 (manual positioning)
%     \draw[->, thick] 
%       (0.5,0) -- (2,0) node[midway, above] {Calibrate};
%   \end{tikzpicture}
%   \vspace{0.5cm} % Vertical space between rows
%   \caption{Illustration of fitting using the convex hull.}
%   \label{fig:bounding_convex_hull}
% \end{figure}

Additionally, computing $C_H$ can be expensive for higher dimensions or for large quantities of data. For these situations, we recommend the first method, which scales very favorably.
\paragraph{Euclidean data-depth for $p_Z$} A fast, but potentially inaccurate, method is to take $p_Z$ to be the data point $X_i$ with the greatest Euclidean depth w.r.t the sample mean $\mu$
\begin{equation*}
  p_{Z} = \arg\max_{X_i} (1 + d_{\text{E}}(X_i))^{-1}
\end{equation*}
where $d_{\text{E}}$ is the Euclidean distance of $x$ to $\mu$
\begin{equation*}
  d_{\text{E}}(x) = \sqrt{(x - \mu)^{\top}(x - \mu)}.
\end{equation*}
\paragraph{Mahalanobis data-depth for $p_Z$} A straightforward improvement is to include the data covariance $\Sigma$ in the depth estimation, using the Mahalanobis distance
\begin{eqnarray*}
  d_{\text{M}}(x) = \sqrt{(x - \mu)^{\top}\Sigma^{-1}(x - \mu)},
\end{eqnarray*}
with again $p_Z$ being taken to be the sample $X_i$ with the greatest depth 
\begin{equation*}
  p_{Z} = \arg\max_{X_i} (1 + d_{\text{M}}(X_i))^{-1} .
\end{equation*}
We find this method also fast, scalable, and more accurate than the Euclidean depth when the data covariance can be inverted.~\cite{messoudi2022ellipsoidal} use a metric similar to $d_{\text{M}}$ as a non-conformity score, the level-sets of which are elliptical.

\paragraph{Approximate Tukey's depth for $p_Z$} A popular measure of depth is Tukey's depth (also known as half-space depth), which for a particular point $X_i$ is defined as the smallest number of points from the dataset that can be contained in any half-spaces passing through $X_i$. I.e. what is the smallest data partition that can be obtained
\begin{equation*}
  d_{T}(x) = \inf_{v\in \mathbb{R}^{d}} \frac{1}{n} \sum^{n}_{j=1} \mathbb{I}\{v^{\top} (X_j - x ) \geq 0\},
\end{equation*}
where $\mathbb{I}$ is the indicator function that the sample $X_i$ is in the half-space defined by vector $v$ and point $x$. We then pick $p_Z$ to be the point with the greatest depth
\begin{equation*}
    p_Z = \arg\max_{X_i} d_{T}(X_i).
\end{equation*} 
Although Tukey's depth is robust, it is highly expensive to compute (requiring 2 loops over the data), we therefore find this approximate Tukey's depth performs quite well up to moderate dimensions
\begin{equation*}
  \tilde{d}_{T}(x) = \inf_{v\in V} \frac{1}{n} \sum^{n}_{j=1} \mathbb{I}\{v^{\top} (X_j - x ) \geq 0\},
\end{equation*}
where are $V$ are the normal vectors of the data enclosing set. That is, one only checks the half-spaces composing the enclosing zonotope $\zonotope$.

\subsection{Calibrating zonotope prediction sets}\label{sec:calibration}

\paragraph{Calibration with a known distribution} For mostly illustrative purposes, we begin by showing how $Z^{\alpha}_{p_Z}$ can be calibrated from a known multivariate distribution, whose cdf $F_X$ or density $f_X$ are available. This could perhaps be useful for some applications, but the next method (calibrating from sample data) is likely to find wider use. Given a zonotope $\zonotope$ (ideally containing the range of $f_{X}$ or otherwise some large probability region ($\mathbb{P}_X(\zonotope)\approx 1$ if $X$ is unbounded) and a $p_Z$ ideally near the mode, the structure can be calibrated by integrating the density $f_X$ in the sets $\nestedzonotope$:
\begin{equation*}
  s(\alpha) = 1 - \int_{\mathcal{Z}^{\alpha}_{p_Z}} f_X(x)dx.
\end{equation*}
The condition 
\begin{equation*}
  \mathbb{P}(X \in \mathcal{Z}^{s(\alpha)}_{p_Z}) \geq 1 - \alpha
\end{equation*}
holds straightaway. 

\paragraph{Calibration from sample data} Our method is similar to conformal prediction, as described in section~\ref{sec:conformal_prediction}, however without a trained regressor $\hat{f}$ and our data $\{X_1, X_2, \dots, X_n\}$ is multidimensional. We also have a parametric set family $\nestedzonotope$ in contrast to a non-conformity scoring function. However, the set-membership of the data $\alpha_i = \sup \{\alpha \in [0, 1] \; \mid \; X_i \in \nestedzonotope\}$ can be used to rank or score the data. Under the assumption of exchangeability, the probability of another $X_{n+1}$ having a membership score as extreme as the $\alpha_i$'s is equation~\ref{eq:pvalues}. This directly leads to the (conservative) quantile $s(\epsilon) = \alpha_{\ceil{\epsilon n}}$ with sorted $\alpha$ producing a set with the property
\begin{equation}\label{eq:zonotope_prob_bound}
  \mathbb{P}(X_{n+1} \in \mathcal{Z}_{p_Z}^{s(\epsilon)}) \geq 1 - \epsilon,
\end{equation}
where $\epsilon \in [0,1]$ is a user defined confidence value.
\begin{figure}[t!]
  \centering
  \includegraphics[width=0.35\textwidth]{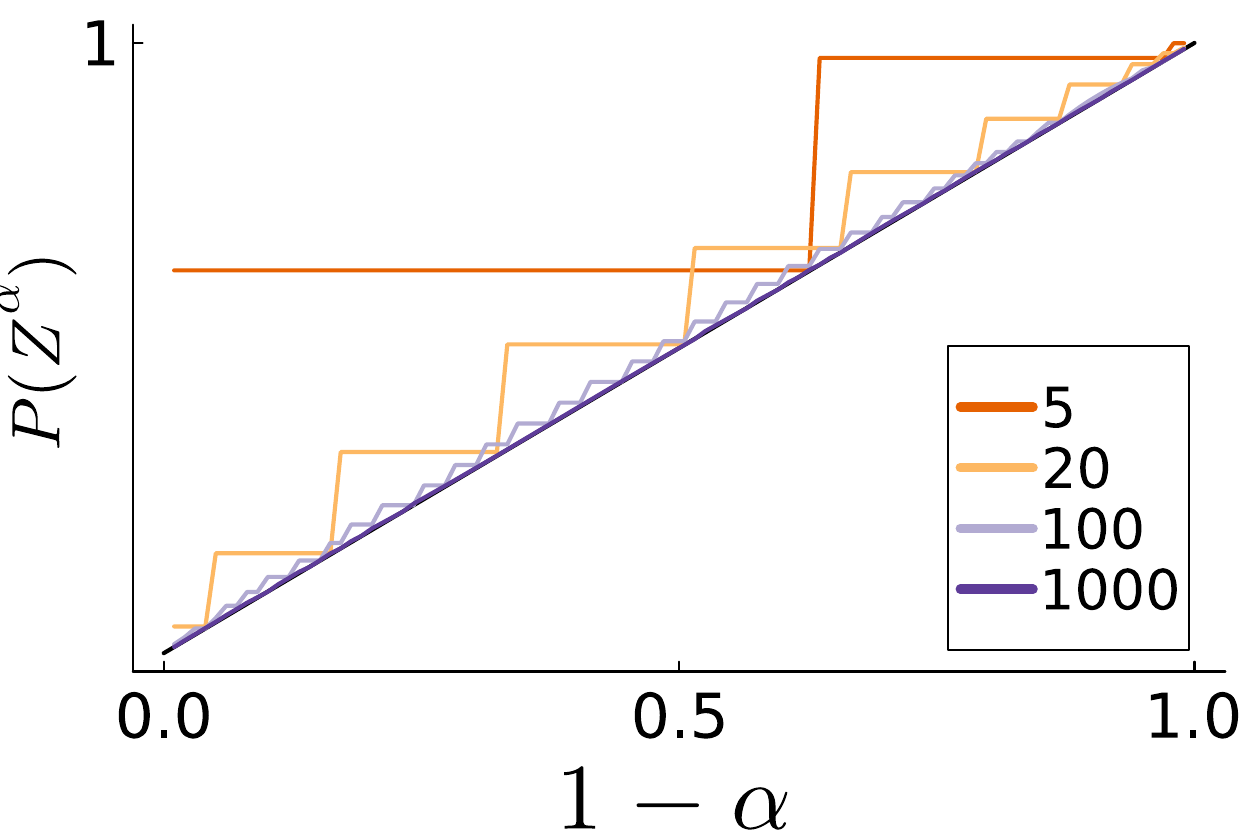}
  \caption{Effect of grid size in membership computation (equation~\ref{eq:membership}). The example computes the prediction sets of a 2-dimensional standard normal gaussian. A large 2M calibration points and 200M points for empirical coverage $\mathbb{P}(\mathcal{Z}^{\alpha})$ are used to isolate the effect of the discretisation.}
  \label{fig:discretisation}
\end{figure}

\begin{figure}[h!]
  \centering
  \begin{subfigure}[b]{0.156\textwidth}
    \centering
    \includegraphics[width=\textwidth]{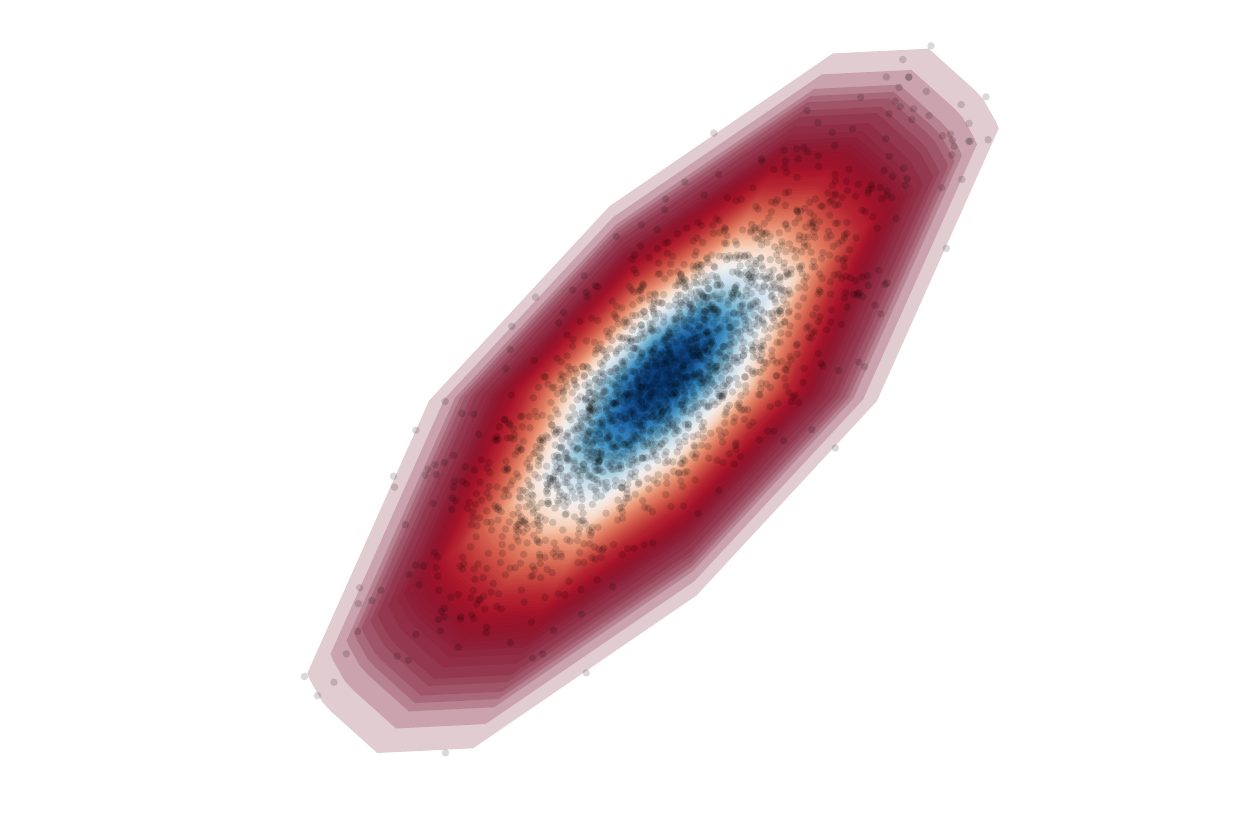}
  \end{subfigure}
  \begin{subfigure}[b]{0.156\textwidth}
    \centering
    \includegraphics[width=\textwidth]{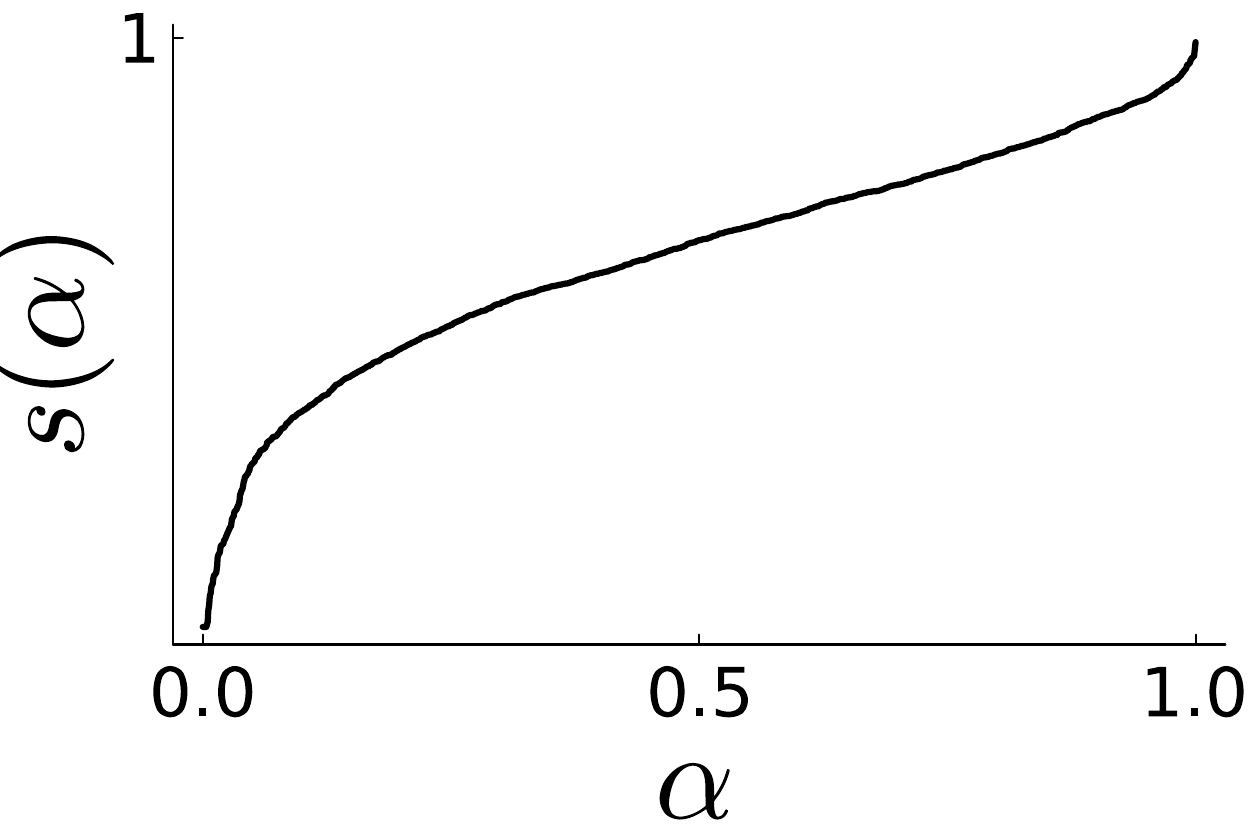}
  \end{subfigure}
  \begin{subfigure}[b]{0.156\textwidth}
    \centering
    \includegraphics[width=\textwidth]{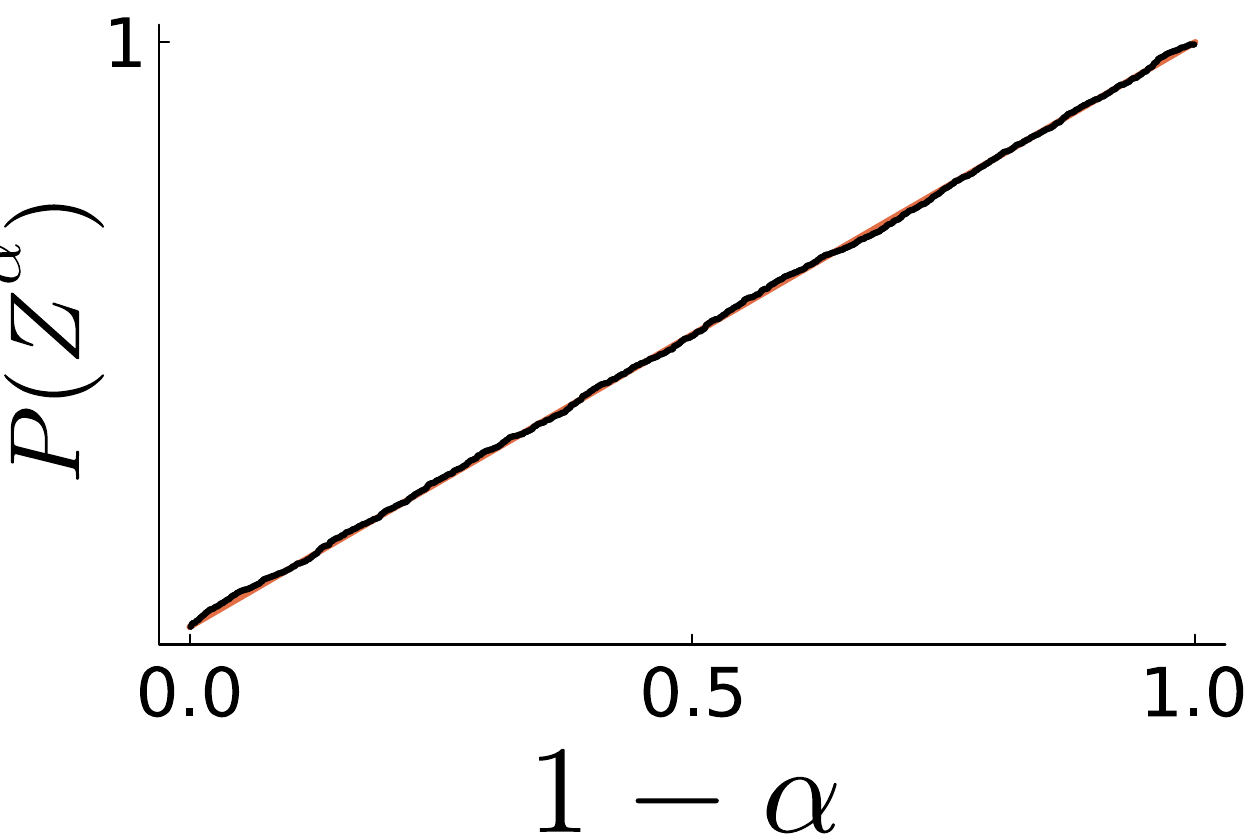}
  \end{subfigure}
  \begin{subfigure}[b]{0.156\textwidth}
    \centering
    \includegraphics[width=\textwidth]{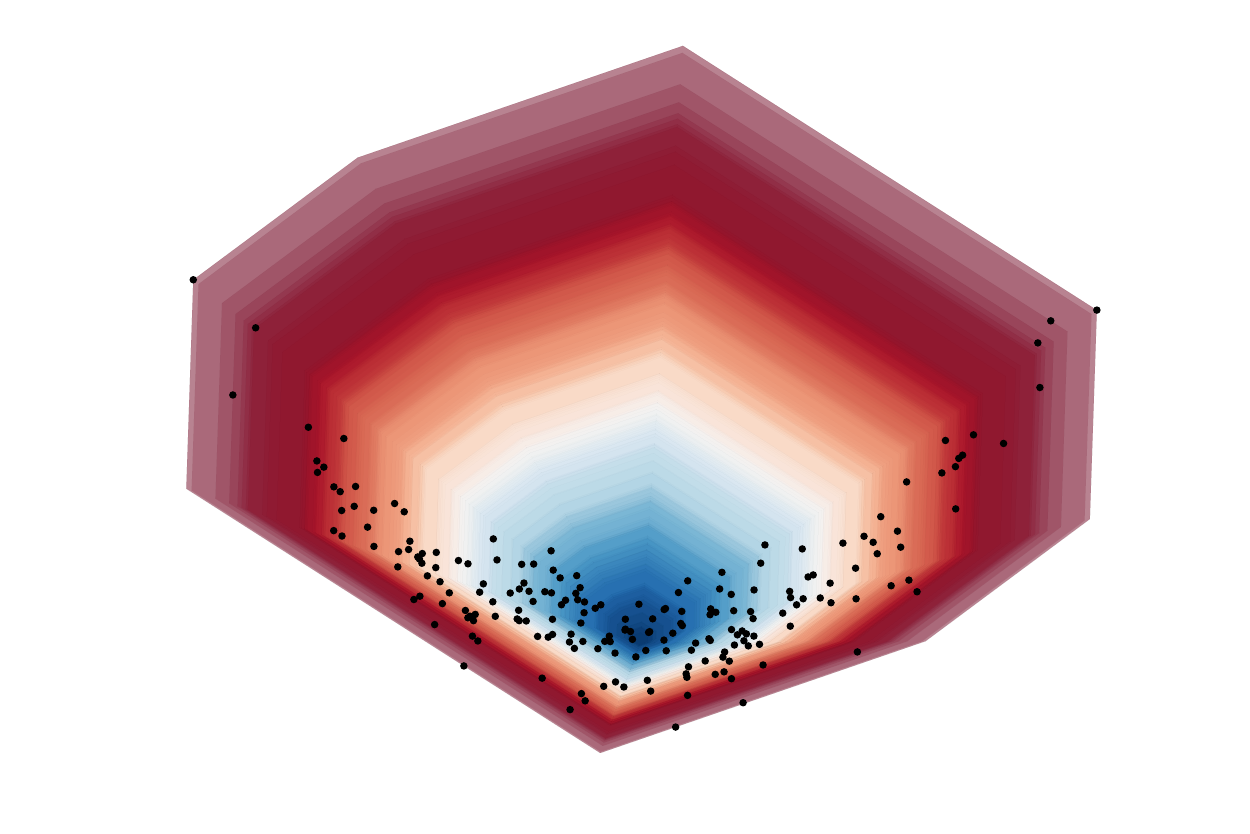}
  \end{subfigure}
  \begin{subfigure}[b]{0.156\textwidth}
    \centering
    \includegraphics[width=\textwidth]{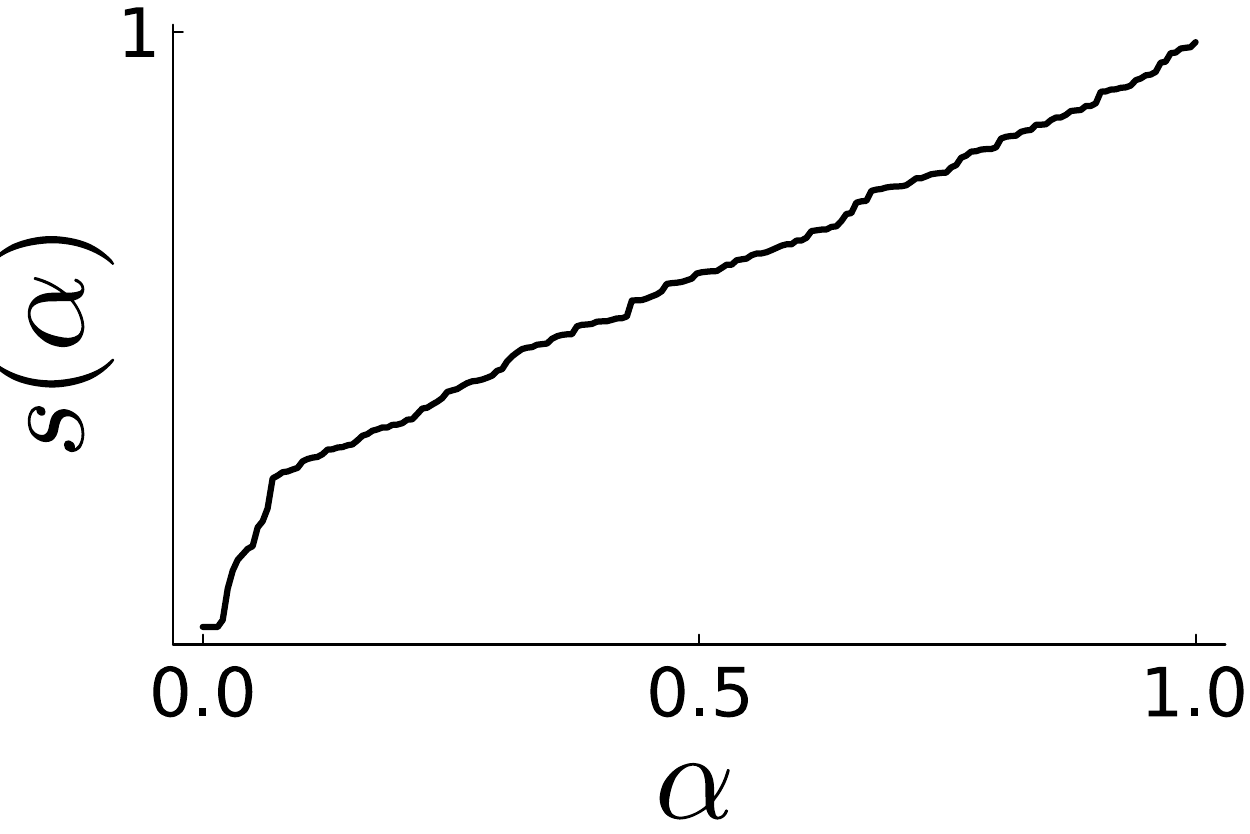}
  \end{subfigure}
  \begin{subfigure}[b]{0.156\textwidth}
    \centering
    \includegraphics[width=\textwidth]{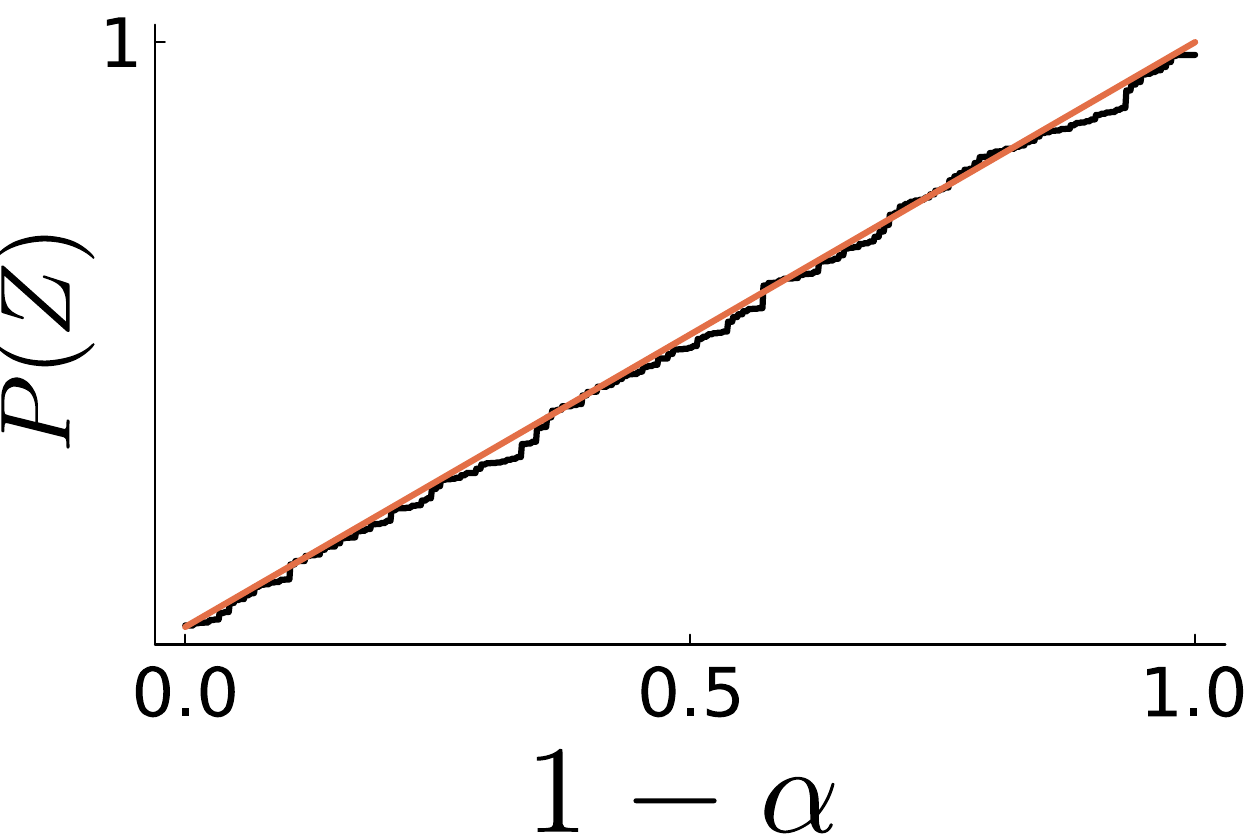}
  \end{subfigure}
  \begin{subfigure}[b]{0.156\textwidth}
    \centering
    \includegraphics[width=\textwidth]{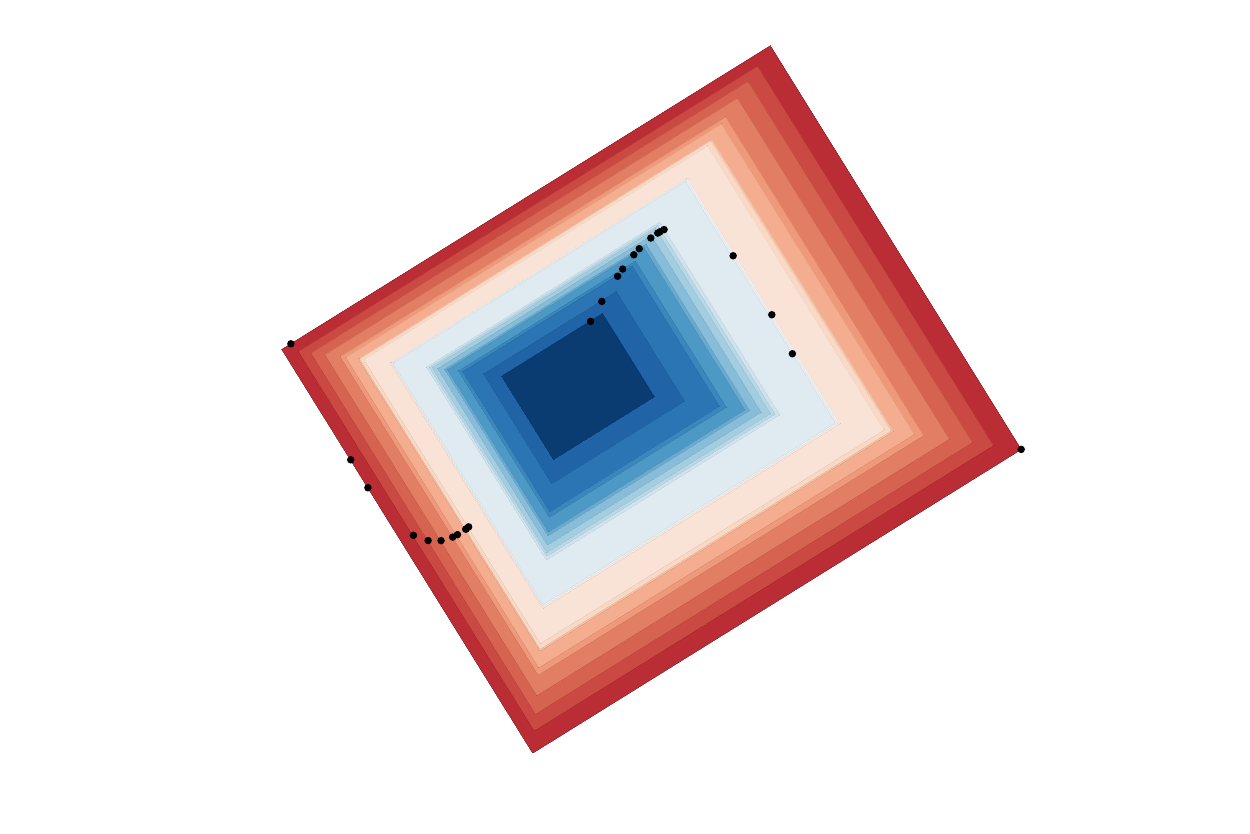}
  \end{subfigure}
  \begin{subfigure}[b]{0.156\textwidth}
    \centering
    \includegraphics[width=\textwidth]{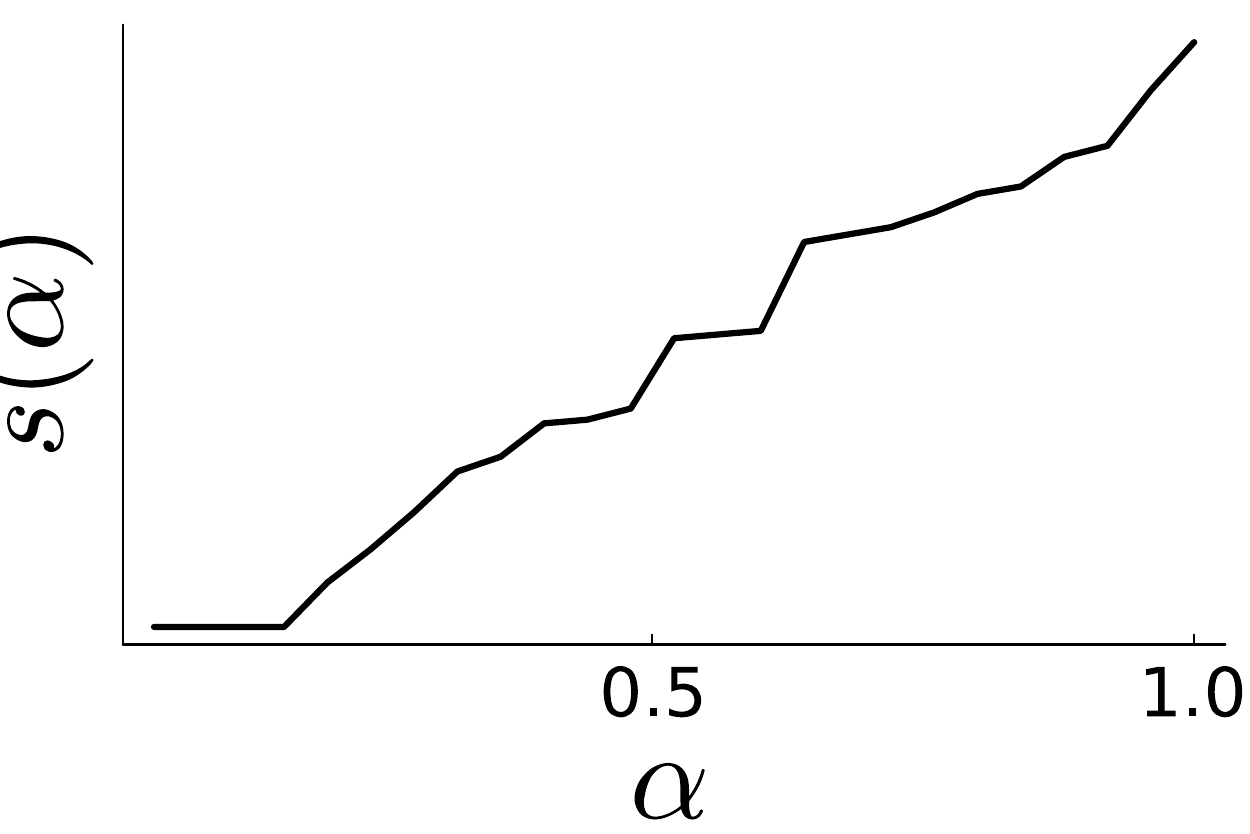}
  \end{subfigure}
  \begin{subfigure}[b]{0.156\textwidth}
    \centering
    \includegraphics[width=\textwidth]{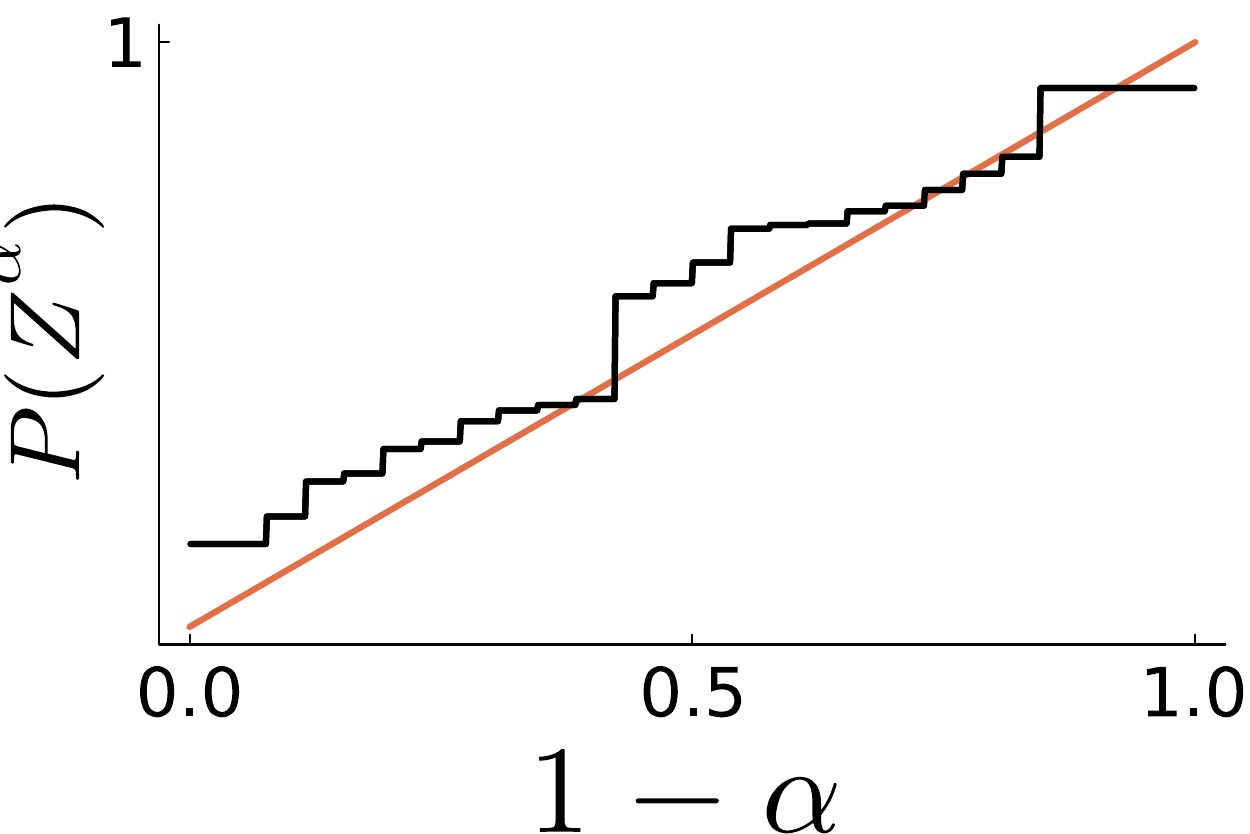}
  \end{subfigure}
  % Adding manually positioned arrows using TikZ
  \caption{Shows three examples of calibration using conformal prediction. The top row shows a correlated Gaussian of $2000$ points, and the middle a $200$ samples of a skewed half-moon, and bottom $25$ samples of a $\sin$ function. The final column shows the empirically tested coverage.}
  \label{fig:calibrated_zonotopes}
\end{figure}

We note that computing the exact membership score $\alpha_i = \sup \{\alpha \in [0,1] \; \mid \; X_i \in \nestedzonotope\}$ may be challenging, as the set must be varied with $\alpha$ until $X_i \notin \nestedzonotope$, which could be relatively well performed using mathematical optimisation for certain set representations. We however suggest a simple method, where $\alpha$ is uniformly discretised in a grid in $A = \{0,0.111, \dots, 0.999, 1\}$, and computing
\begin{equation}\label{eq:membership}
  \alpha_i = \max \{\alpha \in A \; \mid \; X_i \in \nestedzonotope\} .
\end{equation}
Although the exact supremum isn't found, this still gives conservative results, as two collection of scores  $\alpha_i \leq \alpha_j$ will yield a lower bound on probabilistic bound on \ref{eq:zonotope_prob_bound}. I.e. larger prediction set $\mathcal{Z}_{p_Z}^{\alpha_i} \supseteq \mathcal{Z}_{p_Z}^{\alpha_j}$. This claim is evidenced with a numerical experiment in Figure~\ref{fig:discretisation}, where one can observe that irrespective of the discretisation a guarantee can be obtained, however the ``resolution'' of the sets is effected: the more grid points used, the tighter the bounds are. We suggested that the discretisation of the $\alpha$ values is as least a large as the data set, to avoid multiple repetitions of the same $\alpha_i$ values as much as possible. Also note that like in conformal prediction, the maximum number of unique prediction sets that can be obtained is the number of calibration data points provided. Figure~\ref{fig:calibrated_zonotopes} gives three examples of calibrating $\nestedzonotope$ using this method with different data lengths.

Some potential alternative frameworks exist for reliably calibrating $\nestedzonotope$, including inferential models \citep{martin2015inferential}, confidence-structures \citep{balch2012mathematical}, frequency-calibrated belief functions~\citep{denoeux2018frequency}, and scenario theory~\citep{de2021constructing}. There are likely strong links between these methods and what is proposed here.

\section{Functional prediction sets}
Given a pre-trained model $\hat{f}: X \mapsto F$, which maps from Euclidean space $X \in \mathbb{R}^m$ (space of PDE inputs) to a space of functions $F \in \mathcal{F}$ (PDE solutions), discretised on a fine grid $F_1 = [F_1(y_1), \dots,F_1(y_l)] \in \mathbb{R}^l$, we can compute the error of $\hat{f}$'s error on some unseen data $\{(X_1, F_1), \dots, (X_n, F_n)\}$ with $e_i = F_i - \hat{f}(X_i)$. We may then reduce the dimension of $e_i$ using an SVD: $e = U \Sigma V^{\top}$, where the singular values indicate the variance contribution of each eigenvector to the variance of $e_i$. We may truncate the dimensions of the data, capturing some high (often $99\%$) of the overall variance, and projecting the data $e_i$ onto these remaining dimensions, a low dimension representation of $\hat{f}$'s error: $U_i$. Depending on the size of these dimensions, one of the above fitting methods may be applied to find $\nestedzonotope$. We find the `rotated hyperrectangle' method and the Mahalanobis depth scale well to high dimensions, with the convex hull method being tighter for smaller dimensions. $\nestedzonotope$ may then be calibrated, giving $\mathbb{P}(U_{n+1} \in \mathcal{Z}^{\alpha}) \geq 1 - \alpha$. Mapping $\nestedzonotope$ through the linear transformation back to $\mathbb{R}^{l}$ is straightforward for the underlying zonotope. The probabilistic component $\alpha$ is also straightforward: each zonotope level-set may be propagated independently $g(\nestedzonotope)$ for any function $g$, i.e. perform one transformation for each prediction set, with the $\alpha$-guaranteed being retained by the set. Note we may perform this since we compute a multivariate prediction set, this would not be the case had each dimension been fit independently. 

These nested sets can be added to the prediction of $\hat{f}$, yielding a functional prediction set 
\begin{equation*}
  \mathbb{C}^{\alpha} = \hat{f}(X_{n+1}) + U \Sigma \nestedzonotope.
\end{equation*}
However, the resulting confidence regions are not strictly a guaranteed bound on $(X_{n+1}, F_{n+1})$, since we have truncated some data variance during the fitting. Although these dimensions weakly effect the variance of $F_i$, we cannot strictly claim a guaranteed bound. We therefore propose a quite simple method to account for the uncertainty in these dimensions, without including them in the expensive set calibration.
\subsection{Bounding truncation error}
To maintain rigorous coverage guarantees, we bound the uncertainty in the truncated dimensions by constructing a hyperrectangle $E$ that encloses the projection of all calibration errors onto the discarded SVD modes. Specifically, we compute the element-wise minimum and maximum of the projected errors in the truncated space, forming a bounding box. Upon taking the Cartesian product $\mathcal{R}^{\alpha} = \nestedzonotope \times E$, one obtains a zonotope in high dimensions which contracts in the important directions as $\alpha$ varies, but remains constant in the truncated dimensions. The $\alpha$-guarantee remains unchanged due to this operation because: 1) the number of data points remains unchanged (only their dimension), and 2) the data is totally bounded (and remains so as $\alpha$ changes) in the extra dimensions. Slightly more formally, the indicator function for a Cartesian product is
\begin{equation*}
  \mathbb{I}_{\mathcal{Z}\times E}(x_1, x_2) = \mathbb{I}_{\mathcal{Z}}(x_1)\mathbb{I}_{E}(x_2),
\end{equation*}
and since $E$ is a bounding box for the calibration data, $\mathbb{I}_{E}$ always returns `true' during calibration, and thus would not change the scoring in equation~\ref{eq:membership} had it been included. The truncated dimensions may thus be ignored during calibration. Of course, when empirically testing the coverage of a prediction set constructed this way, both the zonotope membership $\mathbb{I}_{\mathcal{Z}}$ and the bounding box membership $\mathbb{I}_{E}$ must be considered.

Indeed, there is a slight performance-loss due to this (our probabilistic bound becomes looser), however this can be expected to be quite minor, as these truncated modes contribute minimally to the overall variance of $F_i$. Our final prediction set becomes
\begin{equation*}
  \mathbb{C}^{\alpha} = \hat{f}(X_{n+1}) + \mathcal{R}^{\alpha},
\end{equation*}
where $\mathcal{R}^{\alpha} = U \Sigma (\nestedzonotope \times E)$ has been projected back.

\paragraph{Cartesian product of zonotopes} The Cartesian product between two zonotopes $\zonotope_X \subset \mathbb{R}^n$ and $\zonotope_Y \subset \mathbb{R}^m$ is a zonotope $\zonotope_{X \times Y} \subset \mathbb{R}^{n\times m}$, whose centre and generator are a simple concatenation of the centres and generators of $\zonotope_X$ and $\zonotope_Y$: $c_{X\times Y} = [c_X\; c_Y]$ and $G_{X\times Y} = [G_X\; G_Y]$.

\subsection{PDE surrogate examples}

We use \texttt{LazySets.jl}~\citep{lazysets21} for the set construction, and \texttt{NeuralOperators.jl}~\citep{pal2023efficient} and some prior models from literature~\citep{gopakumar2024uncertaintyquantificationsurrogatemodels} were used for the base Sci-ML models. 
%Otherwise, a supplementary code repository will be released for those wishing to reproduce this papers results. 
\begin{figure*}[h!]
  \centering
  \begin{subfigure}[b]{0.24\textwidth}
    \centering
    \includegraphics[width=\textwidth]{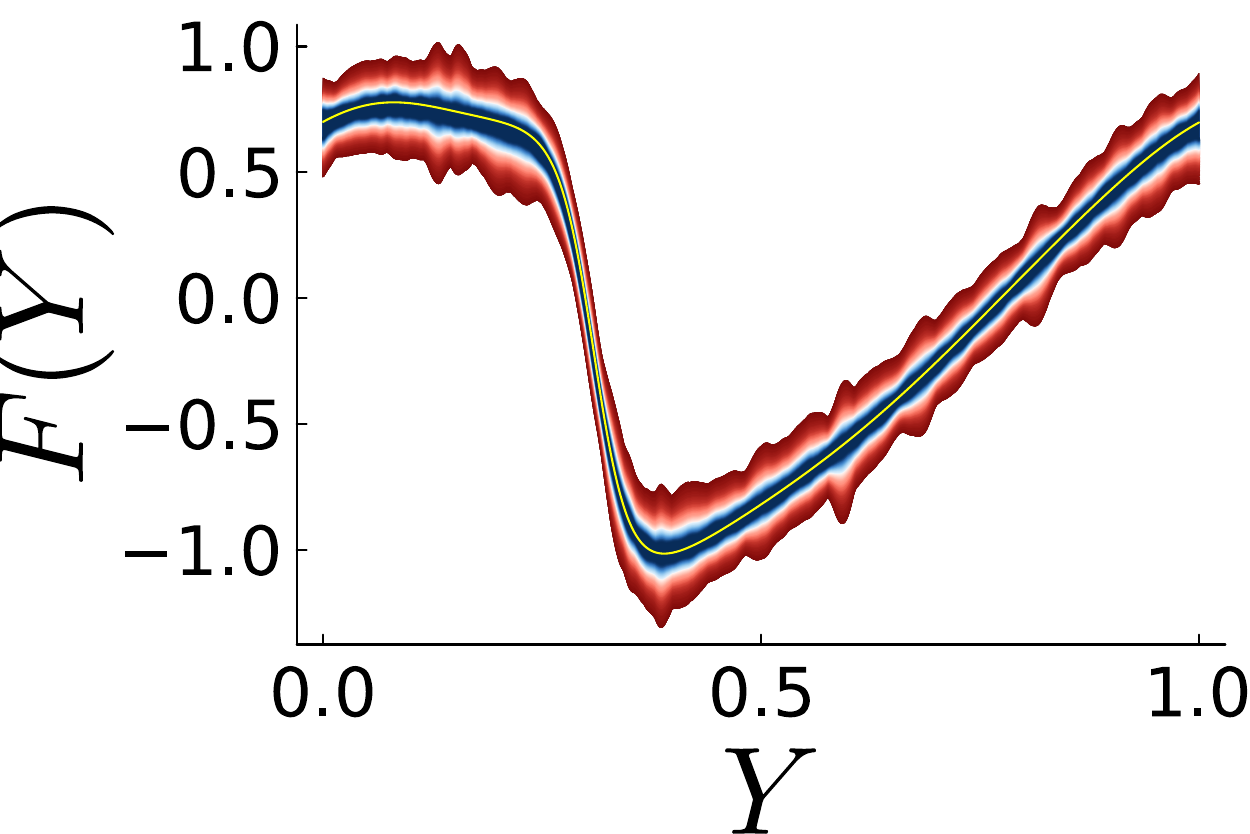}
\end{subfigure}
\begin{subfigure}[b]{0.24\textwidth}
  \centering
  \includegraphics[width=\textwidth]{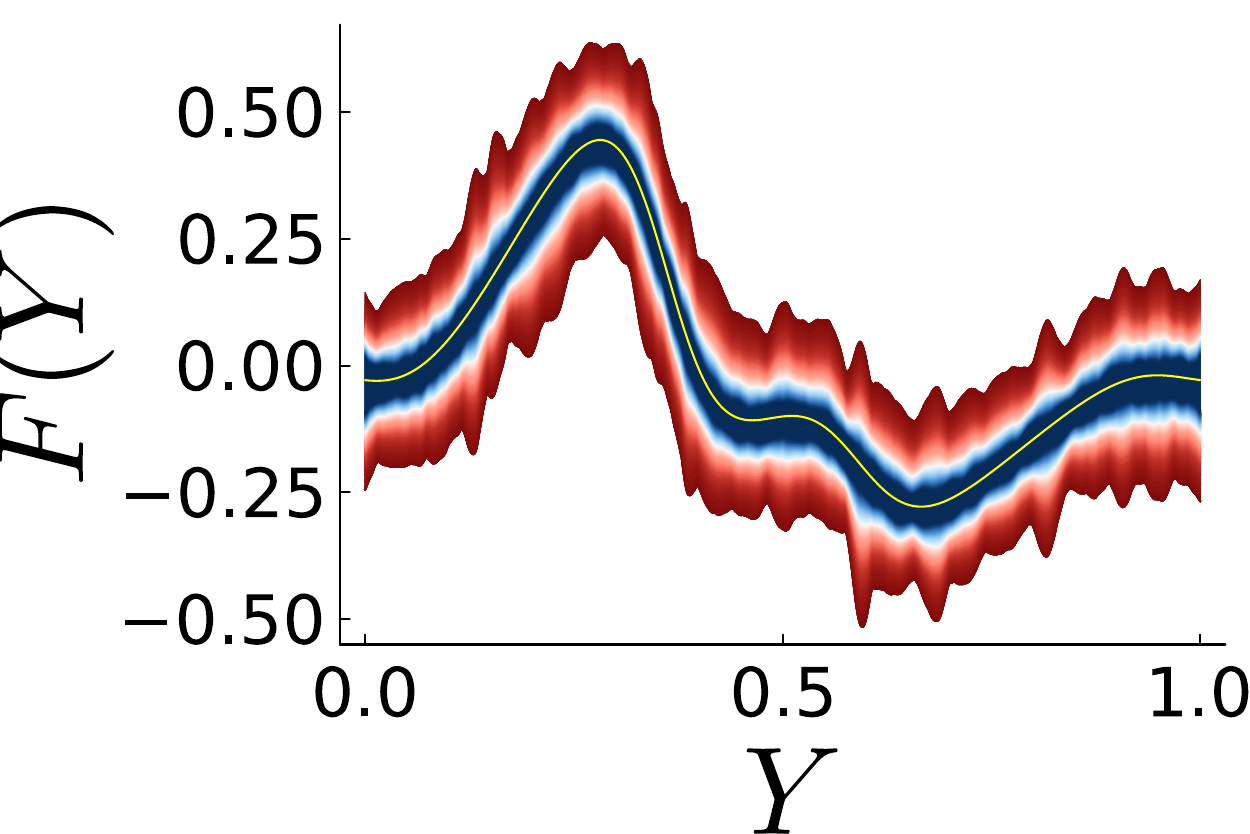}
\end{subfigure}
\begin{subfigure}[b]{0.24\textwidth}
  \centering
  \includegraphics[width=\textwidth]{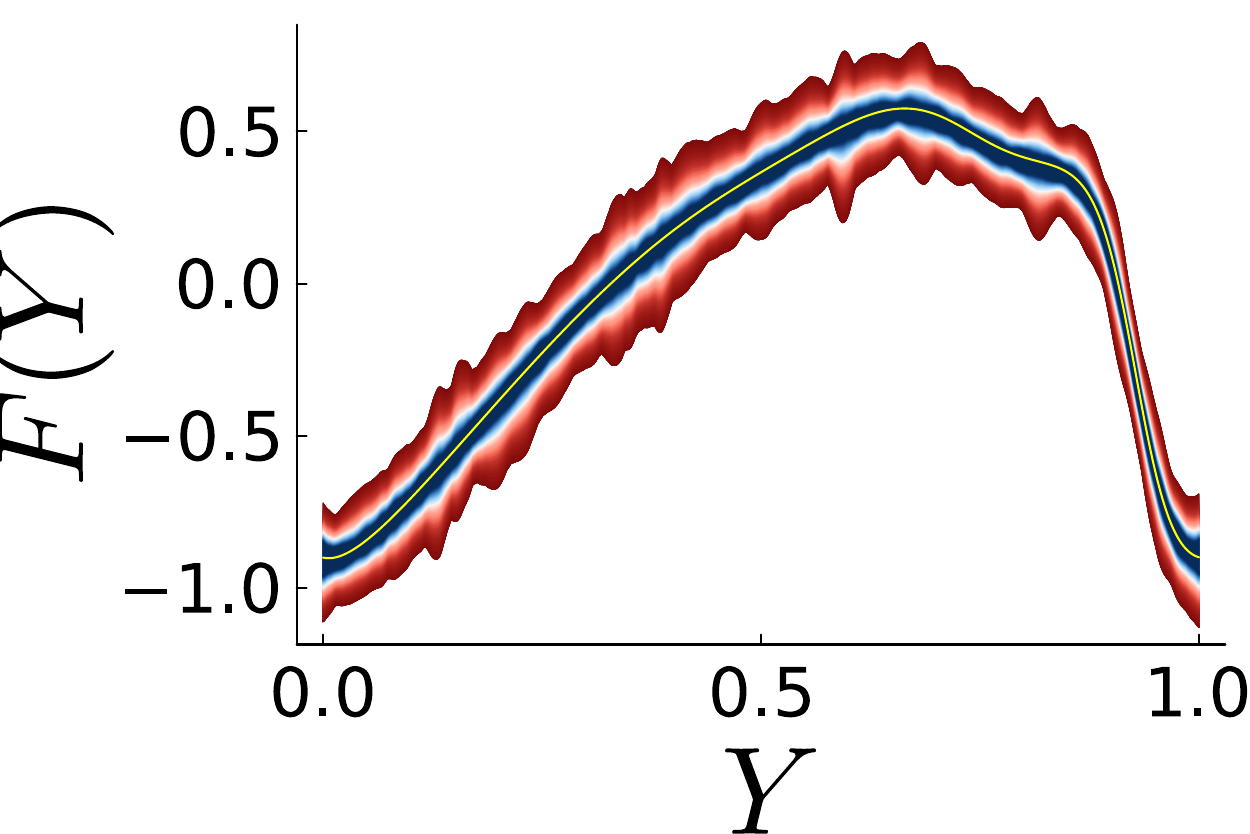}
\end{subfigure}
\begin{subfigure}[b]{0.24\textwidth}
  \centering
  \includegraphics[width=\textwidth]{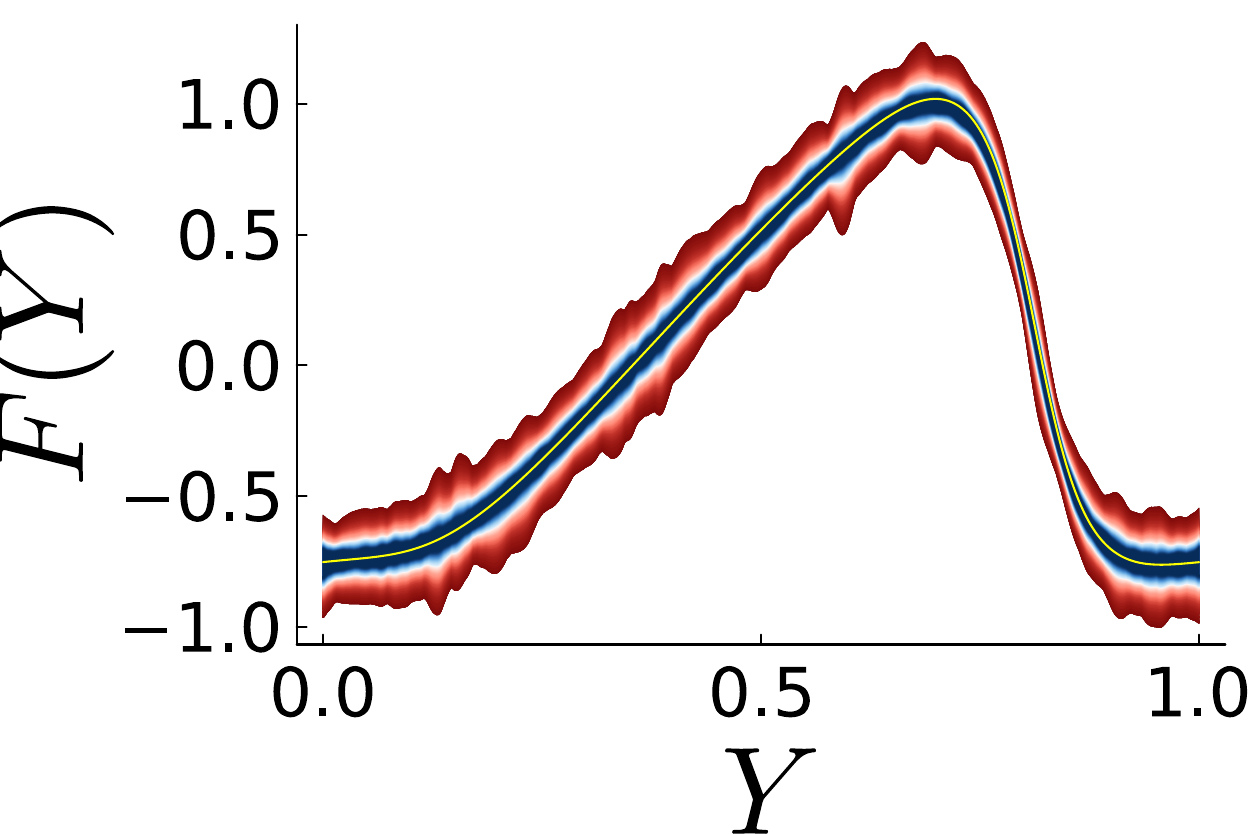}
\end{subfigure}
  \caption{Examples of constructed prediction sets of the FNO for Burger's equation, predicted on test data set. The method predicts nested zonotopes in the high-dimensional space of the surrogate model's prediction. This plot shows the axis-projection of these zonotopes. The ground truth is shown in yellow.}
  \label{fig:FNO_results}
\end{figure*}

To demonstrate the methodology, we construct functional prediction sets on a Fourier neural operator (FNO) for the Burger's equation, trained on data supplementary data provided from~\cite{li2020fourier}, which was built by sampling a numerical PDE solver. After taking an FNO model from literature $1025$ points we used to train the SVD, leaving $500$ for calibrating the prediction sets, and $503$ for validation. Additional information about the training setup and the underlying PDEs are provided in appendix~\ref{app:models}. The predictions from the FNO are discretise onto a grid of 1024 points. A $32$-dimensional rotated hyperrectangle was fitted on the important directions the error's SVD, with $p_Z$ found using the Mahalanobis depth. The entire calibration process took $75$s on a modern laptop, with most of the computation spent on computing the membership~\ref{eq:membership} of the calibration data. Once calibrated, computing a prediction is timed at $0.02$s. Four realisations of the training dataset are shown in Figure~\ref{fig:FNO_results}, with the ground truth shown in yellow.

We also note that a 1024 dimensional nested zonotopic set is predicted for the output of the FNO, what's shown in Figure~\ref{fig:FNO_results} is the axis projection of this set. If we inspect the individual axis, the dependence of the field can also be seen captured.

\begin{table}[t!]
  \centering
  \small 
  \caption{Comparison of empirical coverage and efficiency for various PDEs. Coverage is the empirical frequency with which the true function lies within the predicted set (empirical validation of Eq.~\ref{eq:conformal}). Efficiency is a measured as the volume of the prediction set — smaller values indicate tighter predictions for the same coverage level.}\label{tab:FNO_results}
  \begin{tabular}{rllll}
    \toprule % from booktabs package
       \colorbox{RoyalBlue!0}{$\alpha=0.1$}  &  Modulation  & Rotated  & Zonotope  \\
       \colorbox{Orchid!15}{$\alpha=0.2$}      &           & Box (ours) &  (ours)         \\ 
    \midrule
    PDE (model) &  Coverages   &  & \\
    \midrule % from booktabs package
    Burgers (FNO) &  $89.36$ &  $91.41$ & $88.57$ \\
             &\cellcolor{Orchid!15}$78.32$ &  \cellcolor{Orchid!15}$83.98$ & \cellcolor{Orchid!15}$79$   \\
    Burgers (DeepO) & $87.79$  &$86.8$ & $90.22$  \\
           &\cellcolor{Orchid!15}$76.46$  &\cellcolor{Orchid!15}$79.15$ & \cellcolor{Orchid!15}$82.24$ \\
    Wave (FNO) & $86$ & $89.2$ & $90$  \\
           & \cellcolor{Orchid!15}$78$  & \cellcolor{Orchid!15}$78.4$ &\cellcolor{Orchid!15}$81.6$  \\
    Navier Stokes & $86.83$ & $89.17$ &  $87.33$ \\
     (FNO)    & \cellcolor{Orchid!15}$75.83$  & \cellcolor{Orchid!15}$78.67$ &  \cellcolor{Orchid!15}$79.67$ \\
     \midrule % from booktabs package
       & Efficiency  &  &   \\
    \midrule % from booktabs package
    Burgers (FNO) & $3.956$ & $2.927^{-1}$              & $\mathbf{1.910^{-1}}  $     \\
        & \cellcolor{Orchid!15}$1.634$    & \cellcolor{Orchid!15}$2.562^{-1}$              & \cellcolor{Orchid!15}$\mathbf{1.867^{-1}}  $     \\
    Burgers (DeepO) & $2.742^{2}$      &   $\mathbf{3.79}$          & $5.787$           \\
      &\cellcolor{Orchid!15}$1.049^{2}$ &\cellcolor{Orchid!15}$\mathbf{2.735}$          & \cellcolor{Orchid!15}$5.386$           \\
    Wave (FNO) & $3.588^{4}$& $5.656^{-3}$              &$\mathbf{5.258^{-3}}  $     \\
      & \cellcolor{Orchid!15}$3.559^{4}$         &\cellcolor{Orchid!15}$\mathbf{4.569^{-3}}$          & \cellcolor{Orchid!15}$4.744^{-3}  $         \\
    Navier Stokes  & $9.251^{4}$ & $\mathbf{3.423^{1}}$          & $3.529^{1}  $         \\
    (FNO)    &\cellcolor{Orchid!15}$7.040^{4}$       & \cellcolor{Orchid!15}$\mathbf{2.441^{1}}$          &\cellcolor{Orchid!15}$3.282^{1}  $         \\
    \bottomrule % from booktabs package
  \end{tabular}
  \parbox{0.8\textwidth}{\footnotesize $X^{Y}$ denotes $X\times10^{Y}$}
\end{table}
% \vspace{1cm}
Table~\ref{tab:FNO_results} gives a more in depth comparison of our method to the supremum-based method (labelled \textit{modulation}) from~\cite{diquigiovanni2022conformal}, for various PDEs and models. We note that our method performs favorably in terms of the tightness of the predicted multivariate set (labelled \textit{efficiency}). Standard metrics for efficiency for multivariate conformal prediction are to use the set's volume~\citep{messoudi2022ellipsoidal}. However, since computing the volume of a high dimensional zonotope is challenging, we elect to use the average 2-dimensional projections.
Further details of these benchmarks can be found in appendix~\ref{app:exp} and~\ref{app:models}, including a comparison with lower dimensional data sets from the Mulan and the UCI repository, using more standard MLPs. We note that several other multivariate conformal prediction method are unable to produce results for these models, as they do not scale to the required dimensions. A comparison with these methods is found the appendix in Table~\ref{tab:full_results} for MLPs. We note that for lower dimensional examples, elliptical set conformal prediction performs best, however our method is still competitive.

% \section{Relationships to other ideas}

% \subsection{Non-convex confidence sets}
% It is fairly straightforward to extend the above to other set-representations, for example Taylor-models or polynomial zonotopes, with examples shown in Figure~\ref{fig:nested_topes_poly}. However, a compact representation, like our proposition~\ref{prop:nested_topes} is challenging for more complex set representations. The below sets were constructed by first constructing a $\nestedzonotope \subseteq [-1,1]^p$, and mapping this through the corresponding Taylor polynomial. Although this allows us to compute level-sets (and therefore memberships~\ref{eq:membership}), an efficient method to fit these sets to data would need to be proposed.
% \begin{figure}[h!]
%   \centering
%   \begin{subfigure}[b]{0.156\textwidth}
%     \centering
%     \includegraphics[width=\textwidth]{figures/Confidence_poly_zonotope1.jpg}
% \end{subfigure}
% \begin{subfigure}[b]{0.156\textwidth}
%   \centering
%   \includegraphics[width=\textwidth]{figures/Confidence_poly_zonotope2.jpg}
% \end{subfigure}
% \begin{subfigure}[b]{0.156\textwidth}
%   \centering
%   \includegraphics[width=\textwidth]{figures/Confidence_poly_zonotope3.pdf}
% \end{subfigure}
%   \caption{Example nested polynomial zonotopes.}
%   \label{fig:nested_topes_poly}
% \end{figure}

% This more sophisticated set-representation could allow us to perform a more aggressive non-linear dimension reduction. A potentially interesting application of this work would be to probabilistically fit from data the set-representations used in dynamical system verification. 

\section{Discussion}
% It has also recently been shown by~\cite{balch2019satellite} that Bayesian posteriors (even exact solutions) can suffer from \textit{False Confidence}, meaning that by their nature (additivity) they can produce unsafe results, including in practical/real-world risk calculations, as they demonstrate. Although the advancements in Bayesian machine learning methods has improved AI reliability, for the above reasons alternatives should perhaps be sought for safety-critical systems.
The proposed method is similar to the usual score-based conformal prediction, but where we directly solve for a valid prediction set, instead of taking a level-set of a non-conformity scoring function. Indeed, the membership function of $\nestedzonotope$ is the non-conformity score in conformal prediction~\citep{GUPTA2022108496}. If one could find an efficient (perhaps analytical) expression for this membership function, in terms of $\alpha$ and the underlying set-representation, then the calibration procedure would be made much more efficient (corresponding to a simple function evaluation), without the need to check the set-membership of $\nestedzonotope$ in~\ref{eq:membership}, which is a costly part of the proposition. This would also additionally greatly simplify the implementation.

We therefore share many of the (dis-)advantages of conformal prediction, including finite-sample guarantees and being model agnostic. Although we show example of regression problems, we believe our method (perhaps with a different set-representation) could apply to multivariate classification problems.

\paragraph{Limitations} Since method requires the same assumption as conformal prediction, namely exchangeability, we therefore suffer similar limitations. The guarantee is lost if the dataset is not exchangeable (for example if it changes over time). We also give provide marginal coverage, rather than the stronger conditional coverage, which is more desirable for surrogate modelling. However, methods for improving conditional coverage estimates~\citep{plassier2025rectifying} could be applicable. Other than the usual limitations from conformal prediction, we additionally require an SVD to be trained, which can require a substantial amount of extra training data, in addition to the extra calibration data required for split conformal prediction.

Finally, we mention that our method would greatly benefit for refined methods for fitting zonotopes to data, to improve the tightness of the fitted set.

% \begin{contributions} % will be removed in pdf for initial submission 
% 					  % (without ‘accepted’ option in \documentclass)
%                       % so you can already fill it to test with the
%                       % ‘accepted’ class option
%     Briefly list author contributions. 
%     This is a nice way of making clear who did what and to give proper credit.
%     This section is optional.

%     H.~Q.~Bovik conceived the idea and wrote the paper.
%     Coauthor One created the code.
%     Coauthor Two created the figures.
% \end{contributions}

\begin{acknowledgements} % will be removed in pdf for initial submission,
						 % (without ‘accepted’ option in \documentclass)
                         % so you can already fill it to test with the
                         % ‘accepted’ class option
This project was provided with HPC and AI computing resources and storage by GENCI at IDRIS thanks to the grant 2024-AD010615449 on the \textit{Jean Zay} supercomputer's CSL, V100 and A100 partitions.
\end{acknowledgements}

% References
\bibliography{References}

\newpage

\onecolumn

\title{Guaranteed prediction sets for functional surrogate models\\(Supplementary Material)}
\maketitle

\appendix

\section{Further detail about proposition 1.}
Here we provide additional detail about the proposed nested zonotope family $\mathcal{Z}^{\alpha}_{p_{Z}}$, and a proof that the family is a nested.

\subsection{Intuition behind the nested family}
\begin{figure}[h!]
    \centering
    \includegraphics[width=0.49\textwidth]{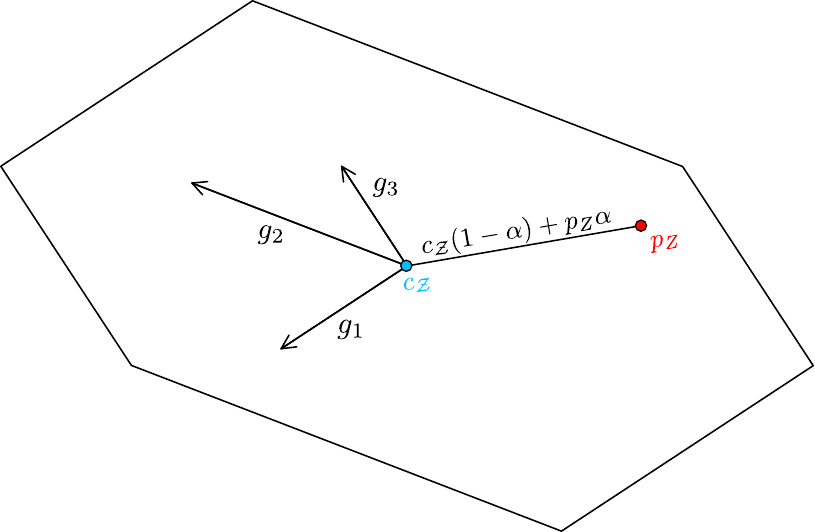}
    \includegraphics[width=0.49\textwidth]{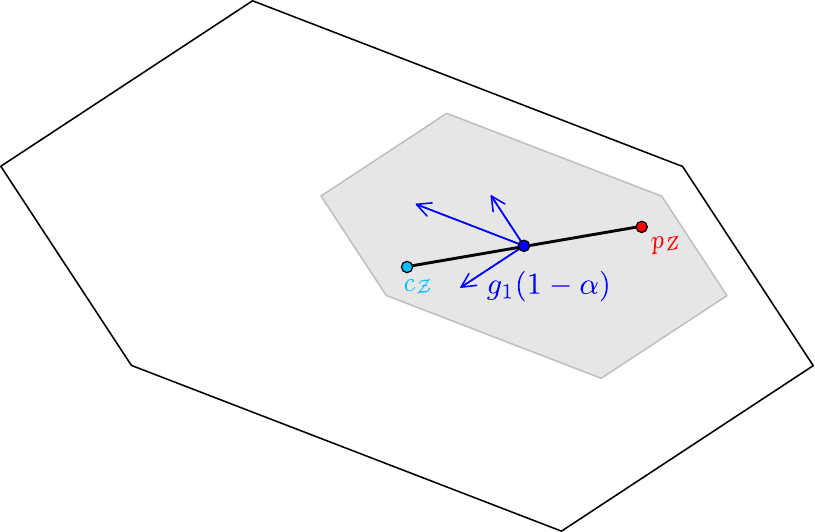}
    \caption{(Left) Shows the outline of an example $\mathcal{Z}$, with generators plotted. Zonotope centre $c_{\mathcal{Z}}$ is in cyan and core $p_Z$ in red, and shows the parametric line defined by $c_{\mathcal{Z}}(1-\alpha) + p_{Z}\alpha$. (Right) Same outline of $\mathcal{Z}$, additionally with the zonotope defined at $\alpha=0.5$ in grey, which has its centre halfway in between $c_{\mathcal{Z}}$ and $p_Z$, and its generators scaled by $0.5$.}
    \label{fig:zonotope_example}
\end{figure}

Figure~\ref{fig:zonotope_example} gives a visual description of how the nested sets $\mathcal{Z}_{p_Z}^{\alpha}$ are constructed. Given a zonotope $\mathcal{Z}$ and a point $p_Z \in \mathcal{Z}$, the centres of the parametric zonotopes move along the line defined by $c_{\mathcal{Z}}(1-\alpha) + p_{Z}\alpha$, and additionally the generators are contracted by $(1-\alpha)$. Thus, the zonotopes are translated toward $p_Z$ as $\alpha$ increases, and their size reduces. Figure~\ref{fig:zonotope_example} shows an example of a parametric zonotope being constructed for $\alpha=0.5$. Note that all the sets in $\mathcal{Z}^{\alpha}_{p_Z}$ have the same shape, and only differ by a translation and a contraction. This is partially because all generators are scaled by the same magnitude. Note: $c_{\mathcal{Z}}$ does not need to be inside all zonotopes, however the point $p_Z$ is inside all zonotopes. Indeed, it is the \textit{only} point that is in all zonotopes (given $\mathcal{Z}^{\alpha=1}_{p_Z} = \{p_Z\})$.

\subsection{Proof of proposition 1.}\label{app:proof}
Here we provide a detailed proof of proposition 1. (nested zonotopic sets), that $\mathcal{Z}^{\alpha_1}_{p_{Z}} \supseteq \mathcal{Z}^{\alpha_2}_{p_{Z}}$ for any $0\leq \alpha_1 \leq \alpha_2 \leq 1$. We remind you that $\mathcal{Z}^{\alpha}_{p_{Z}}$ is defined as
\begin{equation}
    \mathcal{Z}^{\alpha}_{p_{Z}} =\left\{x \in \mathbb{R}^n \Big| \; x = c_{\mathcal{Z}}(1-\alpha) + p_{Z}\alpha + \sum^{p}_{i=1} \xi_ig_i (1- \alpha), \;\; \xi_i \in [-1, 1] \right\},\nonumber
\end{equation}
for some point $p_Z \in \mathcal{Z}$ and $\alpha \in [0,1]$. 
% \begin{proof}
The proof is based on the following rational:
\begin{enumerate}
    \item Take two sets $L$ and $K$ composed as the intersection of $n$ sets $L = \bigcap^{n}_{i=1} A_i$ and $K = \bigcap^{n}_{i=1} B_i$, if each $B_i \subseteq A_i$, then $K \subseteq L$. 
    \item Since a zonotope can be considered as an intersection of half-spaces (H-representation), it suffices to study the subset-hood of the individual half-spaces composing $\mathcal{Z}^{\alpha_1}_{p_{Z}}$ and $\mathcal{Z}^{\alpha_2}_{p_{Z}}$.
    \item A zonotope $\mathcal{Z} = \langle c_{\mathcal{Z}}, G_{\mathcal{Z}}\rangle$ can be characterised as the image of the box $\bigtimes^{n}_{1} [-1, 1]$ in $\mathbb{R}^{n}$, where $n$ is the number generators, by an affine transformation defined by the rotation and stretching with matrix $G_{\mathcal{Z}}$ and translation with vector $c_{\mathcal{Z}}$.
    \item Equivalently it can be defined as the intersection of the same affine transformation of the half-spaces defining the box $\bigtimes^{n}_{1} [-1, 1]$.
    \item A half-space $H_1 = \{x \in \mathbb{R}^n | a_1^{\top}x \leq b_1\}$ is a subset of another half-space $H_2 = \{x \in \mathbb{R}^n | a_2^{\top}x \leq b_2\}$, $H_1 \subseteq H_2$ if:
    \begin{enumerate}
        \item $a_1$ and $a_2$ are positively aligned, i.e., they are parallel and have the same direction: $a_1 \leq \lambda a_2$ for $\lambda \geq 0$.
        \item $b_1 \leq \lambda b_2$, where $\lambda$ the same as (a).
    \end{enumerate}
\end{enumerate}
Condition (a) is straightforward to prove: $\mathcal{Z}^{\alpha_1}_{p_{Z}}$ transforms any half-space normal vectors $a$ as 
\begin{equation}
    (G_{\mathcal{Z}} (1-\alpha_1))^{-\top} a = (1-\alpha_1)^{-1} G_{\mathcal{Z}}^{-\top} a = (1-\alpha_1)^{-1}a',\nonumber
\end{equation}
and $\mathcal{Z}^{\alpha_2}_{p_{Z}}$ as
\begin{equation}
    (G_{\mathcal{Z}} (1-\alpha_2))^{-\top} a = (1-\alpha_2)^{-1} G_{\mathcal{Z}}^{-\top} a = (1-\alpha_2)^{-1}a'. \nonumber
\end{equation}
Therefore, since $\alpha_1 \leq \alpha_2$:
\begin{equation}
    (1-\alpha_1)^{-1}a' \leq (1-\alpha_2)^{-1}a', \nonumber
\end{equation}
holds, and therefore half-spaces transformed by $\mathcal{Z}^{\alpha_1}_{p_{Z}}$ and $\mathcal{Z}^{\alpha_2}_{p_{Z}}$ are parallel and have the same direction. We additionally know that the normal vector $a'$ is contracted by $\lambda = (1-\alpha_2)/ (1-\alpha_1)$ and that $\lambda \in [0, 1]$.
For condition (b), $\mathcal{Z}^{\alpha_1}_{p_{Z}}$ transforms any half-space offset $b$ as 
\begin{align*}
    b_1 &= b + a^{\top}(G_{\mathcal{Z}} (1-\alpha_1))^{-1} (c_{\mathcal{Z}}(1 - \alpha_1) + p_z \alpha_1) \nonumber\\
        &= b + (1-\alpha_1)^{-1}a^{\top}(G_{\mathcal{Z}}^{-1}c_{\mathcal{Z}}(1 - \alpha_1) + G_{\mathcal{Z}}^{-1}p_z \alpha_1).\nonumber
\end{align*}
Setting $J_{\mathcal{Z}} = G_{\mathcal{Z}}^{-1}c_{\mathcal{Z}}$ and $K_{\mathcal{Z}} = G_{\mathcal{Z}}^{-1}p_{Z}$: 
\begin{align}
    b_1 &= b + (1-\alpha_1)^{-1}a^{\top}(J_{\mathcal{Z}}(1 - \alpha_1) + K_{\mathcal{Z}} \alpha_1) \nonumber\\
    &= b + a^{\top}(J_{\mathcal{Z}} + \alpha_1(1-\alpha_1)^{-1}K_{\mathcal{Z}}) \nonumber\\
    &= b + a^{\top}J_{\mathcal{Z}} + \alpha_1(1-\alpha_1)^{-1}a^{\top}K_{\mathcal{Z}}.\nonumber
\end{align}
By symmetry, we also know that for $\mathcal{Z}^{\alpha_2}_{p_{Z}}$:
\begin{equation}
    b_2 = b + a^{\top}J_{\mathcal{Z}} + \alpha_2(1-\alpha_2)^{-1}a^{\top}K_{\mathcal{Z}}.\nonumber
\end{equation}
Inserting these two expressions into our inequality $b_{1} \leq (1-\alpha_2)/ (1-\alpha_1)b_{2}$ and simplifying (noting that $a^{\top}J_{\mathcal{Z}}$ and $a^{\top}K_{\mathcal{Z}}$ are scalar):
\begin{align}
    \alpha_1(1-\alpha_1)^{-1} &\leq  (1-\alpha_2)/ (1-\alpha_1) \alpha_2(1-\alpha_2)^{-1} \nonumber\\
    \alpha_1 &\leq \alpha_2.\nonumber
\end{align}
Thus, any half-space $H$ transformed by $\mathcal{Z}^{\alpha}_{p_z}$ has the property $H^{\alpha_1} \supseteq H^{\alpha_2}$ for any $\alpha_1 \leq \alpha_2$, and $\mathcal{Z}^{\alpha_1}_{p_z} \supseteq \mathcal{Z}^{\alpha_2}_{p_z}$ for any $\alpha_1 \leq \alpha_2$, concluding the proof. 

\section{Relationships between set-representations used in the paper}

We find the \texttt{LazySets.jl}\footnote{https://github.com/JuliaReach/LazySets.jl}~\citep{lazysets21} manual, and software docstrings, quite thorough resources for set-representations and their respective overapproximations. Here we summarise some set isomorphisms and overapproximations used the main paper.

\subsection{Converting hyperrectangles to zonotopes}\label{app:hyperrect}
Hyperrectangles are exactly representable as zonotopes. A hyperrectangle $\mathcal{B} \subset \mathbb{R}^{n}$ with centre vector $C_{\mathcal{B}} \in \mathbb{R}^{n}$ and radius vector $R_{\mathcal{B}}\in \mathbb{R}^{n}$, has the same centre in zonotopic representation $C_{\mathcal{Z}} = C_{\mathcal{B}}$, and a diagonal generator matrix $G_{\mathcal{Z}} \in \mathbb{R}^{n \times n}$ with the radius vector along the diagonals $G_{\mathcal{Z}} = I_n R_{\mathcal{B}}$, where $I_n$ is the identity matrix in $n$ dimensions.

\subsection{Overapproximating a polytope with a zonotope}\label{app:poly}
With the algorithm initially proposed by~\cite{guibas2003zonotopes} (section 4.2), here we summarise the version implemented in \texttt{LazySets.jl}~\citep{lazysets21}. Further detail can be found therein.

Given a polytope $C_{H}$ in vertex representation (for example the result of a convex hull of a dataset) with vertices $v_k$, and some user-selected directions $d_k$ (to which the constructed zonotope's generators will be parallel to), the overapproximation $\mathcal{Z} \supseteq C_{H}$ can be performed by solving the following linear program:
\begin{eqnarray*} 
    \min \sum_{k=1}^l \alpha_k \\
    \text{s.t.} \\
    v_j = c + \sum_{k=1}^l b_{kj} d_k \quad \forall j \\
    -\alpha_k \leq b_{kj} \leq \alpha_k \quad \forall k, j \\
    \alpha_k \geq 0 \quad \forall k. \\
\end{eqnarray*}
The resulting zonotope has center $c$ and generators $\alpha_k d_k$. In this work, we take the directions $d_k$ to be the normal vectors of the enclosing half-spaces the initial polytope.

\section{Extended experiment results}\label{app:exp}
Table~\ref{tab:data_info} gives further information about the data sets, models, and output dimensions for the experiments in this paper. We note that the MLP benchmarks (Bias to Crime) were originally found in~\cite{messoudi2022ellipsoidal} for their comparison of multivariate conformal prediction. Table~\ref{tab:full_results} gives an extended presentation of experimental results, including those for lower dimensional MLPs. Note that for the non-PDE benchmarks, the prediction sets' volume was used as an efficiency metric, while for the high dimensional problem we used average 2D areas, as detailed in the main text.

\begin{table}[t!]
  \centering
  \small 
  \caption{Information about multivariate and PDE datasets}\label{tab:data_info}
  \begin{tabular}{rllll}
    \toprule % from booktabs package
    Name & model  &  Calibration data &  dimensions  & Source  \\
    \midrule
    Bias & MLP& 7750 &  2 & \cite{cho2020comparative}  \\
    Music & MLP& 350  &   2 & \cite{zhou2014predicting}\\
    Indoor & MLP&  6946 &  2 &  \cite{torres2014ujiindoorloc} \\
    SGMM & MLP&  79728 &   4 & \cite{nugteren2015cltune} \\
    Crime & MLP&  628  &  18 &  \cite{redmond2009communities} \\
    Burgers &FNO & 2048  &  1024 & \cite{gopakumar2024uncertaintyquantificationsurrogatemodels} \\
    Burgers &DeepONet &  2048  &  1024 & \cite{gopakumar2024uncertaintyquantificationsurrogatemodels} \\
    Wave & FNO &   7000  & 4096 & \cite{gopakumar2024uncertaintyquantificationsurrogatemodels} \\
    Navier Stokes &FNO & 7000  & 4096 & \cite{gopakumar2024uncertaintyquantificationsurrogatemodels} \\

    \bottomrule % from booktabs package
  \end{tabular}
\end{table}

\begin{table}[t!]
  \centering
  % \small 
  \caption{Results of coverage and efficiency comparison on multivariate and PDE data sets.}\label{tab:full_results}
  \resizebox{\textwidth}{!}{ % Adjust width (0.8) as needed
  \begin{tabular}{rlllll|lllllll}
    \toprule % from booktabs package
   \colorbox{RoyalBlue!0}{$\alpha=0.1$}    &  Coverages  &  &  &  & &  Efficiency  &  &  &  & \\
   \colorbox{Orchid!15}{$\alpha=0.2$}             &  Modulation & Copula & Ellipse & Rotated  & Zonotope  &  Modulation & Copula & Ellipse & Rotated  & Zonotope  \\
    dataset / PDE (model)        &             &        &         & Box (ours) &  (ours) &             &        &         & Box (ours) &  (ours) \\ 
    \midrule % from booktabs package
    Bias & $89.46$ & $89.06$ & $91.13$ & $88.18$ & $91.29$ &  $\mathbf{5.186^{-2}}$  & $5.478^{-2}$     & $5.418^{-2}$     & $5.460^{-2}$              & $6.280^{-2}$ \\
         & \cellcolor{Orchid!15}$78.99$ & \cellcolor{Orchid!15}$79.95$ & \cellcolor{Orchid!15}$81.15$ &  \cellcolor{Orchid!15}$79.15$ & \cellcolor{Orchid!15}$82.59$ & \cellcolor{Orchid!15}$\mathbf{3.346^{-2}}$  & \cellcolor{Orchid!15}$3.508^{-2}$     & \cellcolor{Orchid!15}$3.357^{-2}$     & \cellcolor{Orchid!15}$3.525^{-2}$              & \cellcolor{Orchid!15}$3.913^{-2}$ \\
    Music & $89.71$ & $94.86$ & $96.57$ & $90.86$ & $90.29$ & $\mathbf{4.927^{-1}}$ & $5.504^{-1}$  & $7.467^{-1}    $ & $6.905^{-1}$              & $6.125^{-1}$ \\
          &\cellcolor{Orchid!15}$81.71$ & \cellcolor{Orchid!15}$86.29$ & \cellcolor{Orchid!15}$86.29$ & \cellcolor{Orchid!15}$86.29$ & \cellcolor{Orchid!15}$82.29$ & \cellcolor{Orchid!15}$\mathbf{3.235^{-1}}$ & \cellcolor{Orchid!15}$3.460^{-1}$  & \cellcolor{Orchid!15}$4.104^{-1}    $ & \cellcolor{Orchid!15}$5.512^{-1}$              & \cellcolor{Orchid!15}$4.425^{-1}$ \\
    Indoor & $90.64$ &  $90.96$ &  $90.18$ &  $88.80 $  & $91.22 $ & $8.105^{-2}$     & $8.655^{-2}$  & $\mathbf{7.377^{-2}}$ & $1.217^{-1}$              & $1.377^{-1}$ \\
            & \cellcolor{Orchid!15}$80.33$ &  \cellcolor{Orchid!15}$81.11$ &   \cellcolor{Orchid!15}$81.03$ &  \cellcolor{Orchid!15}$78.72 $  &  \cellcolor{Orchid!15}$81.20 $ & \cellcolor{Orchid!15}$\mathbf{4.494^{-2}}$ & \cellcolor{Orchid!15}$4.580^{-2}$  & \cellcolor{Orchid!15}$4.523^{-2}    $ & \cellcolor{Orchid!15}$7.108^{-2}$              & \cellcolor{Orchid!15}$7.473^{-2}$ \\
    SGMM & $89.98$  & $90.71$ & $89.48$ & $89.88$ & $89.85$ & $2.541^{-5}$     & $2.950^{-5}$  & $\mathbf{1.176^{-9}}$ & $2.224^{-6}$              & $2.632^{-6}$ \\
            & \cellcolor{Orchid!15}$80.01$  & \cellcolor{Orchid!15}$80.93$ & \cellcolor{Orchid!15}$79.37$ & \cellcolor{Orchid!15}$79.74$ & \cellcolor{Orchid!15}$80.28$ & \cellcolor{Orchid!15}$4.142^{-6}$     & \cellcolor{Orchid!15}$4.868^{-6}$  & \cellcolor{Orchid!15}$\mathbf{2.331^{-10}}$ & \cellcolor{Orchid!15}$1.444^{-6}$              & \cellcolor{Orchid!15}$1.715^{-6}$ \\
    Crime & $92.04$ & $89.17$ &$90.45$ & $87.58$ & $91.08$ & $4.125^{-10}$      & $8.923^{-7}$     & $\mathbf{1.583^{-22}}$ & $6.705^{-18}$              & $7.739^{-15}$ \\
           & \cellcolor{Orchid!15}$85.35$ & \cellcolor{Orchid!15}$79.30$ &\cellcolor{Orchid!15}$80.89$ & \cellcolor{Orchid!15}$78.66$ & \cellcolor{Orchid!15}$80.89$ & \cellcolor{Orchid!15}$1.387^{-12}$      & \cellcolor{Orchid!15}$1.889^{-9}$     & \cellcolor{Orchid!15}$\mathbf{4.171^{-25}}$ & \cellcolor{Orchid!15}$8.472^{-19}$              & \cellcolor{Orchid!15}$2.793^{-15}$ \\
    Burgers (FNO) & $89.36$ & ---  &   --- & $91.41$ & $88.57$ & $3.956$      & ---  &   ---       & $2.927^{-1}$              & $\mathbf{1.910^{-1}}  $     \\
            & \cellcolor{Orchid!15}$78.32$ & \cellcolor{Orchid!15}---  &   \cellcolor{Orchid!15}--- & \cellcolor{Orchid!15}$83.98$ & \cellcolor{Orchid!15}$79$ & \cellcolor{Orchid!15}$1.634$      & \cellcolor{Orchid!15}---  &   \cellcolor{Orchid!15}---       & \cellcolor{Orchid!15}$2.562^{-1}$              & \cellcolor{Orchid!15}$\mathbf{1.867^{-1}}  $     \\
    Burgers (DeepO) & $90.92$   &   --- &   --- &$92.77$ & $89.65$ & $2.513^{3}$      &   ---      &   ---       & $\mathbf{5.868}$          & $8.330$           \\
           & \cellcolor{Orchid!15}$80.96$   &   \cellcolor{Orchid!15}--- &   \cellcolor{Orchid!15}--- &\cellcolor{Orchid!15}$83.01$ & \cellcolor{Orchid!15}$79.69$ & \cellcolor{Orchid!15}$9.899^{2}$      &   \cellcolor{Orchid!15}---      &   \cellcolor{Orchid!15}---       & \cellcolor{Orchid!15}$\mathbf{4.052}$          & \cellcolor{Orchid!15}$7.580$           \\
    Wave (FNO) & $86$   & ---   &   ---    & $89.2$ & $90$ & $3.588^{4}$      &   ---      &   ---       & $5.656^{-3}$              & $\mathbf{5.258^{-3}}  $     \\
           & \cellcolor{Orchid!15}$78$   & \cellcolor{Orchid!15}---   &   \cellcolor{Orchid!15}---    & \cellcolor{Orchid!15}$78.4$ & \cellcolor{Orchid!15}$81.6$ & \cellcolor{Orchid!15}$3.559^{4}$      &   \cellcolor{Orchid!15}---      &   \cellcolor{Orchid!15}---       & \cellcolor{Orchid!15}$\mathbf{4.569^{-3}}$          & \cellcolor{Orchid!15}$4.744^{-3}  $         \\
    Navier Stokes & $86.83$ &  --- &  ---  & $89.17$ &   $87.33$ & $9.251^{4}$      &   ---      &   ---       & $\mathbf{3.423^{1}}$          & $3.529^{1}  $         \\
     (FNO)    & \cellcolor{Orchid!15}$75.83$ &  \cellcolor{Orchid!15}--- &  \cellcolor{Orchid!15}---  & \cellcolor{Orchid!15}$78.67$ &   \cellcolor{Orchid!15}$79.67$ & \cellcolor{Orchid!15}$7.040^{4}$      &   \cellcolor{Orchid!15}---      &   \cellcolor{Orchid!15}---       & \cellcolor{Orchid!15}$\mathbf{2.441^{1}}$          & \cellcolor{Orchid!15}$3.282^{1}  $         \\
    \bottomrule % from booktabs package
  \end{tabular}
  }
  \parbox{0.8\textwidth}{\footnotesize $X^{Y}$ denotes $X\times10^{Y}$ and "---" denotes no result}
\end{table}

\section{Physics, Surrogate Models, and Training}\label{app:models}

Here we provide extra details about the PDEs, the functional surrogate models, and their training configurations, used in the main paper. The numerical solvers, data and functional surrogates models used for PDE modelling is borrowed from \citep{gopakumar2024uncertaintyquantificationsurrogatemodels}. 

\subsection{Burgers' Equation}
The Burgers' equation is a partial differential equation often used to model the convection-diffusion of a fluid, gas, or non-linear acoustics. The one-dimensional equation is
\begin{equation}
    \frac{\partial u}{\partial t} + u \frac{\partial u}{\partial y} = \nu\frac{\partial^2 u}{\partial^2 y}, \nonumber
\end{equation}
where $u$ defines the field variables, $\nu$ the kinematic viscosity, and with $y$ and $t$ being the spatial and temporal coordinates respectively. We define a family of initial conditions as follows:
\begin{equation}
    u(y, t = 0) = \sin(\alpha \pi y) + \cos(-\beta \pi y) + \frac{1}{\cosh(\gamma \pi y)}, \nonumber
\end{equation}
parameterised by $\alpha \in [-3, 3]$, $\beta \in [-3, 3]$, and $\gamma \in [-3, 3]$.

\paragraph{Data set generation} A dataset of $2048$ (training) + $1000$ (calibration) + $1048$ (validation) PDE solutions is generated by Latin Hypercube sampling (uniform) the $\alpha$, $\beta$, and $\gamma$ parameters (thus generating random initial conditions for the PDE), and then solving Burgers' 
equation using a spectral solver~\citep{canuto2007spectral}. Each simulation is run for $500$-time iterations with a $\Delta t = 0.0025$ time step and a spatial domain spanning $[0, 1]$, uniformly discretised into $1024$ spatial units. The field at the last time point is then saved as the output, and the surrogate's task is to learn the mapping from the initial condition to the fields final state $u(y, 0) \rightarrow u(y, t_{\text{end}})$.

\subsection{Wave Equation}

\begin{equation}
\frac{\partial^2 u}{\partial t^2} = c^2 \left(\frac{\partial^2 u}{\partial x^2} + \frac{\partial^2 u}{\partial y^2}\right), \nonumber
\end{equation}
where $u$ defines the field variable, $c$ the wave velocity, with $x$, $y$ and $t$ being the spatial and temporal coordinates respectively. The initial conditions are defined as:
\begin{equation}
u(x,y, t = 0) = \exp(-\alpha((x-\beta)^2 + (y-\gamma)^2)), \nonumber
\end{equation}
parameterised by $\alpha \in [10, 50]$, $\beta \in [0.1, 0.5]$, and $\gamma \in [0.1, 0.5]$, with an additional constraint $\frac{\partial u}{\partial t}(x,y,t=0) = 0$.

\paragraph{Data set generation} A dataset of $500$ (training) + $1000$ (calibration) + $1000$ (validation) PDE solutions is generated by Latin Hypercube sampling the $\alpha$, $\beta$, and $\gamma$ parameters. The wave equation is solved using a spectral solver with leapfrog time discretization and Chebyshev spectral method on tensor product grid~\citep{GOPAKUMAR2023100464}. Each simulation runs for $150$ time iterations with $\Delta t = 0.00667$ across a spatial domain of $[-1,1]^2$, discretized into $33$ spatial units per dimension. The first $80$ time instances of each simulation are used for training.

\subsection{Navier-Stokes Equations}
The Navier-Stokes scenario that we are interested in modelling is taken from the exact formulation in \cite{li2020fourier}, where the viscosity of the incompressible fluid in 2D is expressed as:
\begin{align}
    \pdv{w}{t} + u \nabla w  &= \nu  \nabla^2 w + f, &\quad x \in (0,1) , \; y\in (0,1) , \; t\in (0,T)\\
    \nabla u &= 0, &\quad x \in (0,1) , \; y\in (0,1) , \; t\in (0,T)\\
    w &= w_0, &\quad x \in (0,1) , \; y\in (0,1) , \; t=0,
\end{align}
where $u$ is the velocity field and vorticity is the curl of the velocity field $w = \nabla \cross u$. The domain is split across the spatial domain characterised by $x,y$ and the temporal domain $t$. The initial vorticity is given by the field $w_0$. The forcing function is given by $f$ and is a function of the spatial domain in $x,y$. We utilise two datasets from \cite{li2020fourier} that are built by solving the above equations with viscosities $\nu = 1e-3$ and $\nu=1e-4$ under different initial vorticity distributions. For further information on the physics and the data generation, refer \cite{li2020fourier}.

% \subsection{Models}

\subsection{Multilayer Perceptrons}
Multilayer Perceptrons (MLPs), a fundamental class of neural networks, are composed of sequential layers of neurons that transform input features through learned weight matrices and non-linear activations \citep{haykin1994neural}. For a given input $x \in \mathbb{R}^{d_{\text{in}}}$, an MLP with $L$ layers computes:
\begin{align}
h_{i+1} &= \sigma(W_i h_i + b_i), \quad i = 0,\ldots,L-1 \nonumber \\
h_L &= W_L h_L + b_L
\end{align}
where $h_0 = x$, $W_i \in \mathbb{R}^{d_{i+1} \times d_i}$ and $b_i \in \mathbb{R}^{d_{i+1}}$ are learnable parameters, and $\sigma$ is a non-linear activation function (in this case, hyperbolic tangent).

\subsection{Fourier Neural Operators}

Fourier Neural Operators (FNOs), introduced by~\cite{li2020fourier}, are specific instance of the Neural Operator (NO) class of ML models, which have shown efficacy in mapping between function spaces. Following a description by~\cite{gopakumar2024plasma}, a NO can be written as a parameterised mapping between function spaces $G_{\theta}: A \mapsto U$, where $G_{\theta}$ is a neural network parameterised by $\theta$, with three specific architecture elements, which are sequential:
\begin{enumerate}
    \item \textbf{Lifting}: A fully local, point-wise operation that projects the input domain to a higher dimensional latent representation $a \in \mathbb{R}^{d_a} \rightarrow \nu_0 \in \mathbb{R}^{d_{\nu_0}}$,
    \item \textbf{Iterative Kernel Integration}: Expressed as a sum of a local linear operator, and a non-local integral kernel operator, that iterates $\nu_i \rightarrow \nu_{i+1}$ for several layers:
    \begin{equation*}
        \nu_{i+1} = \sigma(W \nu_i(x) + \kappa(a; \phi)\nu_i(x)),
    \end{equation*}
    where $W$ is the learnable linear components and $\sigma$ is a non-linear activation (as in traditional networks). The kernel $\kappa$ (with learnable parameters $\phi$) characterises a neural network's layer as a convolution, as the following integral with the prior layer's output $\nu_i$:
    \begin{equation}
        (\kappa(a; \phi)\nu_{i})(x) = \int_{D} \kappa(x, y, a(x), a(y); \phi)\nu_i(y)dy. \nonumber
    \end{equation}
    \item \textbf{Projection} Similar to lifting, but with reversed dimensions $\nu_n \in \mathbb{R}^{d_{\nu_n}} \rightarrow u \in \mathbb{R}^{d_{u}}$, where $n$ is the total number of layers.
\end{enumerate}

A Fourier Neural Operator is a specific class of the above, which defines $\kappa$ with Fourier convolutions:
\begin{equation}
    (\kappa(a; \phi)\nu_i)(x) = \mathcal{F}^{-1}(\mathcal{F}(R_{\phi}) \cdot \mathcal{F}(\nu_i)), \nonumber
\end{equation}
where $\mathcal{F}$ and $\mathcal{F}^{-1}$ are the Fourier and inverse Fourier transform, and $R_{\phi}$ is a learnable complex-valued tensor comprising of truncated Fourier modes. In practical application, the network is discretised to a finite number of modes and the discrete FFT is used for $\mathcal{F}$. FNOs are learnable using gradient-based optimisation using automatic differentiation.

\subsection{MLP Training}
The trained MLP model consists of:
\begin{enumerate}
\item \textbf{Input layer}: $\mathbb{R}^{d_{\text{in}}} \rightarrow \mathbb{R}^{256}$
\item \textbf{Hidden layers}: 3 layers of $\mathbb{R}^{256} \rightarrow \mathbb{R}^{256}$ transformations with tanh activation
\item \textbf{Output layer}: $\mathbb{R}^{256} \rightarrow \mathbb{R}^{d_{\text{out}}}$
\end{enumerate}
The network was trained using the Adam optimizer with an initial learning rate of $5 \times 10^{-3}$, which was reduced by a factor of 0.5 every 50 epochs. Training proceeded for 500 epochs using mini-batches of size 50 and mean squared error loss. Input features were normalized using a min-max scaling strategy applied per variable. The total parameter count for a single input/output dimension is $256(d_{\text{in}} + 1) + 256^2(L-1) + 256 + d_{\text{out}}(256 + 1)$.

\subsection{FNO Training}
The trained FNO model consists of
\begin{enumerate}
    \item \textbf{Lifting layer}: a full connected layer $\mathbb{R}^2 \rightarrow \mathbb{R}^{64}$ ($192$ parameters)
    \item \textbf{Fourier layers}: 4 Fourier layers ($69696$ parameters each), each consisting of:
    \begin{enumerate}
        \item a full connected later $\mathbb{R}^{64} \rightarrow \mathbb{R}^{64}$ ($4160$ parameters)
        \item Fourier operator kernel $\mathbb{R}^{64} \rightarrow \mathbb{R}^{64}$, truncated to $16$ modes ($65536$ parameters)
    \end{enumerate}
    \item \textbf{Projection layer}: two layer fully connected network $\mathbb{R}^{64} \rightarrow \mathbb{R}^{125} \rightarrow \mathbb{R}^{1}$, with Gelu (Gaussian Error Linear Unit) activation (8449 paramaters).
\end{enumerate}
The FNO was trained for 500 epochs using the Adams optimiser with stepsize of $10^{-3}$ and a weight decay of $10^{-4}$, on an $L_2$ loss, and is timed at around 8 minutes on an NVIDIA A100 GPU.

\end{document}